\documentclass{article}

\PassOptionsToPackage{numbers,compress}{natbib}
\usepackage[preprint]{neurips_2026}

\usepackage[utf8]{inputenc} % allow utf-8 input
\usepackage[T1]{fontenc}    % use 8-bit T1 fonts
\usepackage{hyperref}       % hyperlinks
\usepackage{url}            % simple URL typesetting
\usepackage{booktabs}       % professional-quality tables
\usepackage{graphicx}       % figures and \resizebox
\usepackage{amsmath}        % advanced math
\usepackage{amsfonts}       % blackboard math symbols
\usepackage{amssymb}        % symbols like \checkmark
\usepackage{multirow}       % multi-row table cells
\usepackage{float}          % [H] float placement
\usepackage{nicefrac}       % compact symbols for 1/2, etc.
\usepackage{array}          % custom column types
\usepackage{microtype}      % microtypography
\usepackage[table]{xcolor}  % colors and colored tables
\usepackage{wrapfig}        % text-wrapping figures and tables
\usepackage{tabularx}       % appendix-wide tables
\newcommand{\best}[1]{\textbf{#1}}

\newcommand{\sidetablecaption}[2]{\refstepcounter{table}{Table \thetable:} #2\par\label{#1}\vspace{0.35em}}
\newcommand{\equalcontrib}{\ensuremath{\dagger}}

\title{EventPrune: Cascaded Event-Assisted Token Pruning for Efficient First-Person Dynamic Spatial Reasoning}

\author{%
\textbf{Pengtao Ma}\textsuperscript{1,\equalcontrib},
\textbf{Ziliang Zhou}\textsuperscript{1,\equalcontrib},
\textbf{Ciyu Ruan}\textsuperscript{1},
\textbf{Haoyang Wang}\textsuperscript{1},
\textbf{Kaiyuan Li}\textsuperscript{1},
\\
\textbf{Zihang Gong}\textsuperscript{2},
\textbf{Wenhua Ding}\textsuperscript{1},
\textbf{Chen Gao}\textsuperscript{3},
\textbf{Jingao Xu}\textsuperscript{4},
\textbf{Xinlei Chen}\textsuperscript{1}
\\[0.5em]
\textsuperscript{1}Shenzhen International Graduate School, Tsinghua University,
\textsuperscript{2}Harbin Institute of Technology,
\\
\textsuperscript{3}Tsinghua University,
\textsuperscript{4}The University of Hong Kong
}

\begin{document}

\maketitle

\begingroup
\renewcommand{\thefootnote}{\equalcontrib}
\footnotetext{Equal contribution.}
\endgroup

\begin{abstract}

First-person dynamic spatial reasoning requires models to track continuous motion and precise geometric structure, but the quadratic attention cost of Transformer-based Video-LLMs makes dense visual tokens computationally expensive. 
Existing token pruning paradigms predominantly rely on discrete static snapshots, failing to preserve the motion and geometric cues essential for reasoning.
We propose Event Cascade Pruning (ECP), to our knowledge the first training-free framework that leverages the high-frequency motion cues from event cameras as a continuous event-guided motion prior to guide token selection.
ECP combines three stages: Event-Triggered Causal Sampling to anchor motion-informative keyframes, Event-guided Motion Saliency Filtering to suppress event-inactive visual tokens, and Event-Attention Ranking Fusion to calibrate spatial attention with motion-salient dynamics. 
With 80\% visual token reduction, ECP outperforms the full-token baseline (37.62\% vs. 36.31\%) while achieving \(1.89\times\) inference speedup and 52\% GFLOPs reduction. 
We further introduce ESR-Real, the first real-world RGB-event benchmark for first-person spatial reasoning, where ECP improves accuracy by 2.68 percentage points over full-token baselines.

\end{abstract}

\section{Introduction}

% Video-LLMs~\cite{llava_onevision, longvu, qwen2_5_vl} are shifting video understanding from passive captioning to active first-person dynamic spatial reasoning~\cite{vsibench, rem_spatial2024, rynnec2025}. 
% Recent works have begun integrating these models into embodied intelligence to handle complex tasks like indoor navigation and drone perception~\cite{uni_navid2024, convoi2024, embodied_videoagent2024}. 
% Unlike traditional video understanding, embodied scenarios are characterized by intense ego-motion from a first-person perspective~\cite{eagle2024, alanavlm2024}, which introduces significant spatiotemporal dynamics. 
% These dynamics demand that models not only process information efficiently~\cite{prunevid2024, videollm_mod2024} but also maintain precise geometric structures and motion continuity to make accurate physical judgments.

Video-LLMs~\citep{llava_onevision, longvu, qwen2_5_vl} are increasingly used for first-person dynamic spatial reasoning~\citep{vsibench, rem_spatial2024, rynnec2025},
such as indoor navigation and drone perception~\citep{uni_navid2024, convoi2024, embodied_videoagent2024}.
Unlike conventional video understanding, embodied scenarios involve rapid egocentric motion and continuously changing scene geometry~\citep{eagle2024, alanavlm2024}.
These settings require models to process video efficiently~\citep{prunevid2024, videollm_mod2024}
while preserving motion continuity and geometric structures for reliable spatial reasoning.

% To mitigate the ``visual token explosion'' and high inference latency in these real-time scenarios, training-free visual token pruning has emerged as a critical optimization strategy~\cite{fastv, pyramiddrop, prunevid2024}. However, the high-dynamic nature of embodied video poses fundamental challenges for existing software-based pruning paradigms at both inter-frame and intra-frame levels:
% (1) For inter-frame pruning, existing methods filter temporal redundancy by comparing pixel differences or feature similarities~\cite{dtd}. Such appearance-based comparison not only introduces significant latency but also fails to capture continuous dynamic characteristics like speed and trajectory, making it unreliable for \textbf{dynamic information} extraction.
% (2) For intra-frame pruning, existing methods rely on either attention weights~\cite{fastv, vtw, pyramiddrop} or diversity metrics~\cite{divprune, btp}. Our analysis reveals a severe spatial bias in attention scores: weights concentrate disproportionately on image borders (Fig.~\ref{fig:spatial_bias_combined}), causing erroneous elimination of central regions; diversity-based strategies, while aiming to maximize visual coverage, forcibly merge spatially adjacent tokens and sever local geometric continuity, fragmenting object boundaries and contours that are crucial for physical judgments. 
% In short, existing intra-frame pruning methods fail to preserve \textbf{geometric structures} essential for spatial reasoning.

To mitigate the ``visual token explosion'' and high inference latency in these
real-time settings, training-free visual token pruning has become an important
optimization strategy~\citep{fastv, pyramiddrop, prunevid2024}. However,
existing RGB-only pruning signals are fragile for highly dynamic egocentric
videos at both temporal and spatial levels.
(1) For inter-frame pruning, existing methods filter temporal redundancy by comparing pixel differences or feature similarities~\cite{dtd}. Such appearance-based comparison not only introduces significant latency but also fails to capture continuous dynamic characteristics like speed and trajectory, making it unreliable for dynamic information extraction.
(2) For intra-frame pruning, methods based on attention weights~\citep{fastv, vtw, pyramiddrop}
or diversity metrics~\citep{divprune, btp} may discard structure-critical
tokens. Our analysis reveals a severe peripheral spatial bias in attention
scores: attention concentrates disproportionately on image borders
(Fig.~\ref{fig:spatial_bias_combined}), causing task-relevant interior regions
to be under-retained. Diversity- or merging-based strategies may scatter retained
tokens or merge local neighborhoods, fragmenting object boundaries and weakening
geometric continuity. Consequently, existing pruning methods often fail to
preserve the motion and structural cues required for spatial reasoning under
high compression.

\begin{wrapfigure}{r}{0.50\columnwidth}
    \centering
    \includegraphics[width=\linewidth]{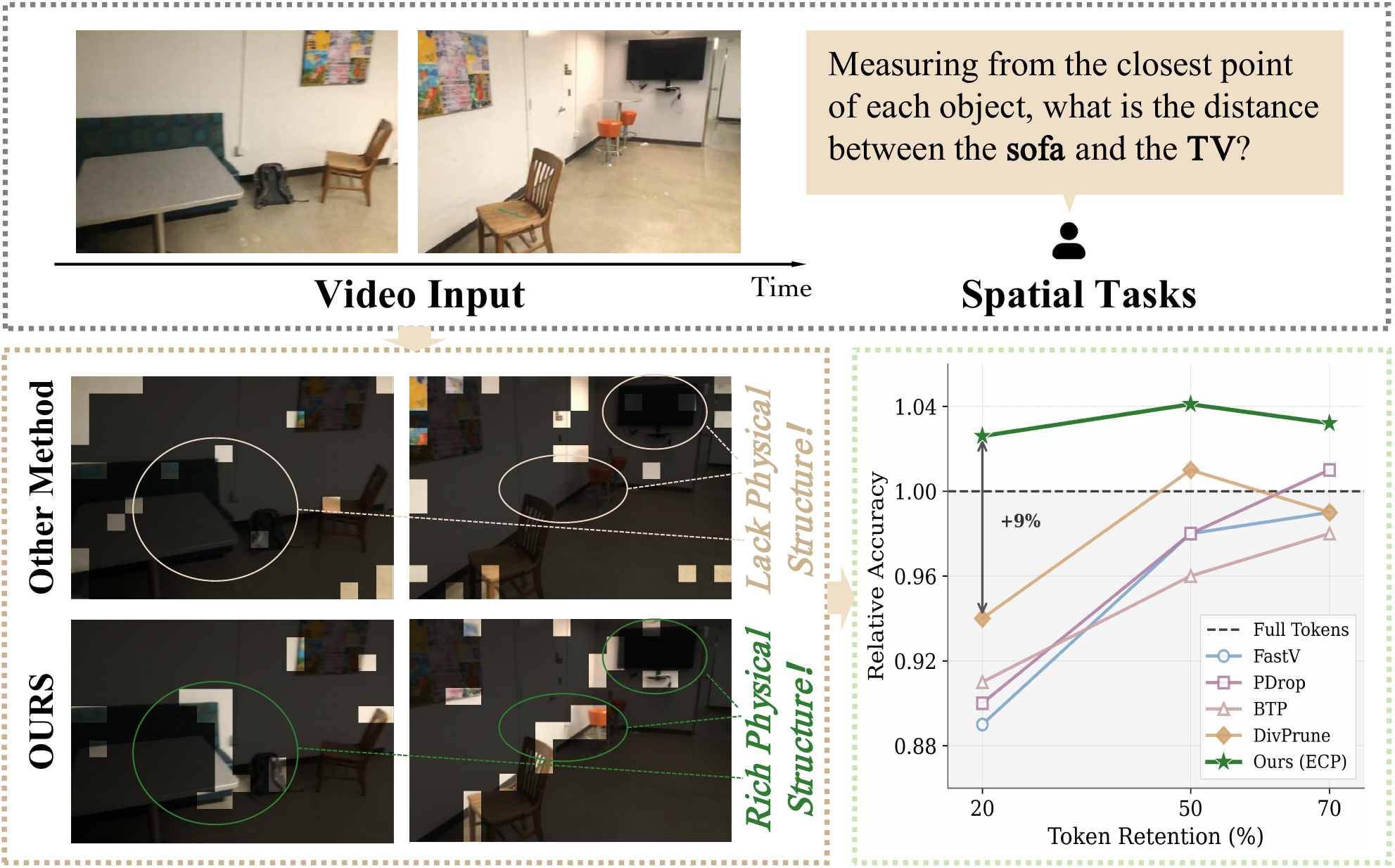}
     \vspace{-0.4cm} % 稍微收缩标题与正文之间的垂直距离
    % \caption{\textbf{Comparison of token pruning results for spatial reasoning tasks.} Existing pruning methods discard structural cues such as object boundaries, while ECP preserves them even at high sparsity.}
\caption{Token pruning for spatial reasoning.
Unlike prior pruning, ECP preserves task-relevant geometric cues under high sparsity.}
\vspace{-0.4cm}
    \label{first pic}
\end{wrapfigure}

In contrast to existing software-based approaches, we identify event cameras as a natural external cue for these issues. As bio-inspired sensors that asynchronously record pixel-level brightness changes with microsecond resolution, event cameras offer unique cues for both dynamics and geometry:
(1) At the inter-frame level, event streams directly record motion activity at each pixel. This provides an instant, training-free prior for anchoring causally critical transitions and filtering temporally redundant regions without expensive feature matching.
(2) At the intra-frame level, events exhibit high sensitivity to edges and textures, representing the geometric structures that biased attention mechanisms neglect. These signals offer a reliable reference to calibrate attention scores and ensure the retention of tokens with true spatial saliency.

% However, events are asynchronous external signals, fundamentally heterogeneous from the synchronous grid features inside Video-LLMs. How to deeply integrate this heterogeneous prior into the inference pipeline to genuinely guide token selection remains non-trivial.
% To overcome this, we propose the Event Cascade Pruning (ECP) framework, a training-free paradigm establishing a full-stage physical-semantic joint mechanism.
% ECP comprises three cascaded modules: (1) Event-Triggered Causal Sampling (ETCS) anchors keyframes via physical activity flux and temporal gradients; (2) Event-guided Motion Saliency Filtering (EMSF) constructs a lightweight dynamic filter to prune low-dynamic regions; and (3) Event-Attention Ranking Fusion (EARF) calibrates the model's spatial attention bias through non-parametric rank alignment, preserving tokens with dual physical-semantic saliency.

However, events are asynchronous external signals, fundamentally heterogeneous from the synchronous grid features inside Video-LLMs. How to deeply integrate this heterogeneous prior into the inference pipeline to genuinely guide token selection remains non-trivial.
To overcome this, we propose Event Cascade Pruning (ECP), a training-free event-assisted
pruning framework with three cascaded modules. Event-Triggered Causal Sampling
(ETCS) anchors keyframes using event activity flux and temporal changes;
Event-guided Motion Saliency Filtering (EMSF) removes low-motion regions via
event saliency maps; and Event-Attention Ranking Fusion (EARF) calibrates
attention-based token scores through non-parametric rank alignment, preserving
tokens with joint event-semantic saliency.

% Extensive experiments demonstrate the effectiveness of ECP. Under 80\% token reduction, ECP surpasses all existing pruning methods and outperforms the full-token baseline (37.62\% vs. 36.31\%), achieving 1.89$\times$ speedup and 52\% GFLOPs reduction. On our real-world ESR-Real benchmark with native event signals, ECP further improves over baseline by 2.68\% at 50\% retention. These results validate that event-assisted pruning simultaneously enhances both efficiency and accuracy for first-person dynamic spatial reasoning.

Extensive experiments demonstrate the effectiveness of ECP. At 20\% token retention, ECP outperforms all evaluated training-free pruning
baselines and surpasses the full-token baseline (37.62\% vs. 36.31\%), while
achieving 1.89$\times$ speedup and 52\% GFLOPs reduction. On our real-world
ESR-Real benchmark with native event signals, ECP improves over the full-token
baseline by 2.68 points at 50\% retention. These results show that
event-assisted pruning improves efficiency while preserving, and
in some cases improving, accuracy for first-person dynamic spatial reasoning.

% Our main contributions are: 
% \textbf{(1)} We propose Event Cascade Pruning (ECP), the first training-free event-driven pruning framework for first-person dynamic spatial reasoning, establishing a cascaded temporal-to-spatial filtering pipeline that preserves both dynamics and geometry.
% \textbf{(2)} We introduce ESR-Real, the first real-world RGB-event benchmark for spatial reasoning with over 700 QA pairs across 6 categories.
% \textbf{(3)} We achieve state-of-the-art results: ECP outperforms full-token baselines at 80\% compression with 1.89$\times$ speedup and 52\% GFLOPs reduction, with consistent gains on both simulated and real-world benchmarks.

Our main contributions are:

\textbf{(1)} We propose Event Cascade Pruning (ECP), to our knowledge the first
training-free event-assisted pruning framework for first-person dynamic
spatial reasoning, establishing a cascaded temporal--spatial--representation pruning
pipeline that removes redundancy while preserving motion, geometric, and
attention-derived semantic cues.

\textbf{(2)} We introduce ESR-Real, a real-world RGB-event benchmark for
first-person spatial reasoning with over 700 QA pairs across 6 categories.

\textbf{(3)} We achieve the best performance among evaluated training-free
pruning methods: ECP outperforms the full-token baseline at 80\% token reduction
with 1.89$\times$ speedup and 52\% GFLOPs reduction, with consistent gains on
both simulated-event and real-event benchmarks.

\section{Related Work}
\label{sec:related_work}

\begin{figure*}[h]
  \centering
  % 0.9\linewidth 表示占页面总宽度的 90%
  \includegraphics[width=0.95\linewidth]{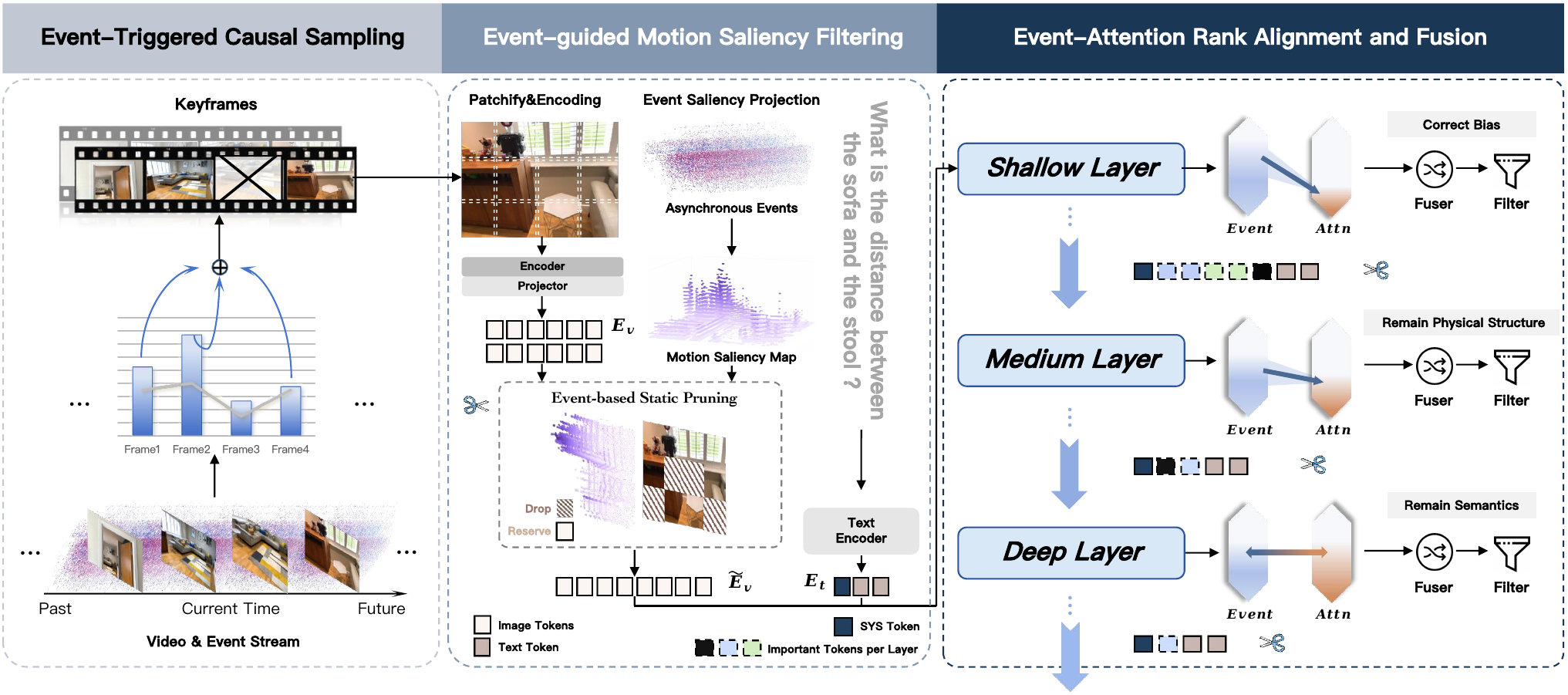}
  \vspace{-0.2cm} % 稍微收缩图片与标题之间的垂直距离，可根据需要调整
  \caption{
Overview of Event Cascade Pruning (ECP). ECP comprises three cascaded modules: ETCS selects keyframes via event activity flux; EMSF prunes static regions using motion saliency maps; EARF calibrates attention with event priors, applying layer-wise fusion weights.
}
  \label{fig:framework}
  \vspace{-0.3cm} % 稍微收缩标题与正文之间的垂直距离
\end{figure*}

% \textbf{Video-LLMs and First-Person Dynamic Spatial Perception.}
% Recent Large Vision-Language Models (LVLMs)~\cite{flamingo, blip2, llava, qwen_vl, cogvlm, internvl} integrate visual encoders with LLMs~\cite{llama2, gpt4} to achieve unified multimodal understanding. To capture fine-grained details, representative models like LLaVA-NeXT~\cite{llava_next} and Qwen2.5-VL~\cite{qwen2_5_vl} adopt dynamic high-resolution strategies. While effective for traditional image understanding tasks~\cite{video_llama, videochatgpt}, extending these architectures to continuous sequences~\cite{llava_onevision, longva} causes a visual token explosion. This computational burden prohibits efficient first-person dynamic spatial reasoning, which demands rapid responsiveness to continuous physical dynamics (e.g., velocity)---a requirement unmatched by current heavy-weight architectures.

\textbf{Video-LLMs and First-Person Dynamic Spatial Perception.}
Recent Large Vision-Language Models (LVLMs)~\cite{flamingo, blip2, llava, qwen3_vl, cogvlm, internvl}
integrate visual encoders with LLMs~\cite{llama2, gpt4} for unified multimodal
understanding. Representative Video-LLMs such as LLaVA-NeXT~\cite{llava_next},
Qwen2.5-VL~\cite{qwen2_5_vl}, and LLaVA-OneVision~\cite{llava_onevision} improve fine-grained and long-context video understanding,
but their high-resolution visual tokens lead to substantial computational cost.
This bottleneck is especially problematic for first-person dynamic spatial
reasoning, where models must respond to rapid ego-motion and changing geometry
under latency constraints.

% \textbf{Visual Token Pruning.}
% To mitigate computational bottlenecks, visual token pruning has emerged as a critical optimization strategy~\cite{sparsevit, llmlingua}, operating primarily across inter-frame and intra-frame dimensions. 
% At the inter-frame level, traditional methods and the recent DTD~\cite{dtd} rely on comparing inter-frame pixel or feature differences to discard redundant tokens.
% At the intra-frame level, approaches like FastV~\cite{fastv}, VTW~\cite{vtw}, and PyramidDrop~\cite{pyramiddrop} select regions based on attention scores, while ToMe~\cite{tome} merges tokens via bipartite matching. DivPrune~\cite{divprune} further introduces diversity-based optimization. Recently, BTP~\cite{btp} proposed balancing local saliency and global diversity.
% However, these paradigms face severe challenges in first-person dynamic spatial reasoning: appearance-based metrics overlook continuous physical dynamics; attention mechanisms disproportionately concentrate weights on image borders and corners~\cite{xiao2024efficient}; and forced diversity selection disrupts local geometric continuity. Consequently, they fail to maintain the integrity of both semantic understanding and physical structure at high compression rates.

\textbf{Visual Token Pruning.}
% Training-free visual token pruning reduces this cost along temporal and spatial
% dimensions~\cite{sparsevit, llmlingua}.
To mitigate computational bottlenecks, visual token pruning has emerged as a critical optimization strategy~\cite{sparsevit, llmlingua}, operating primarily across inter-frame and intra-frame dimensions. 
Inter-frame methods such as DTD~\cite{dtd}
remove redundancy via pixel or feature differences, while intra-frame methods
select or merge tokens using attention~\cite{fastv, vtw, pyramiddrop},
diversity~\cite{divprune}, token merging~\cite{tome}, or hybrid
objectives~\cite{btp}. However, these software-only cues are fragile for
first-person dynamic spatial reasoning: appearance differences miss
high-frequency motion continuity, attention scores suffer from peripheral
spatial bias~\cite{xiao2024efficient}, and diversity or merging objectives can
fragment local geometric structures. Consequently, existing pruning methods
often discard motion- or structure-critical tokens under high compression.

% \textbf{Event-based Vision in VLMs.}  Event cameras (Dynamic Vision Sensors) asynchronously capture pixel-level brightness changes with microsecond resolution, offering intrinsic high dynamic range and motion sensitivity~\cite{gallego2020event, chakravarthi2024event_survey, wang2025event_mobile}. With recent commercialization by manufacturers such as Prophesee and Sony, these sensors are gaining traction in embodied AI applications including autonomous navigation and robotics. While demonstrating exceptional performance in vision tasks such as SLAM~\cite{eventslam, zhu2019event}, optical flow~\cite{eventflow}, and video reconstruction~\cite{vid2e, e2vid}, their integration into LVLMs remains limited. Recent efforts such as EventCLIP~\cite{eventclip} and EventBind~\cite{eventbind} align event representations with CLIP embeddings for recognition, while EventGPT~\cite{eventgpt} extends this paradigm to event captioning. However, these approaches require training dedicated event encoders and cannot leverage pretrained RGB-based VLMs. In contrast, we propose a training-free framework that utilizes events as a lightweight event-guided motion prior to guide token pruning in existing Video-LLMs, preserving dynamic and geometric information without parameter updates.

\textbf{Event-based Vision in VLMs.}  Event cameras (Dynamic Vision Sensors) asynchronously capture pixel-level brightness changes with microsecond resolution, offering intrinsic high dynamic range and motion sensitivity~\cite{gallego2020event, chakravarthi2024event_survey, wang2025event_mobile}. With recent commercialization by manufacturers such as Prophesee and Sony, these sensors are gaining traction in embodied AI applications including autonomous navigation and robotics. While demonstrating exceptional performance in vision tasks such as SLAM~\cite{eventslam, zhu2019event}, optical flow~\cite{eventflow}, and video reconstruction~\cite{vid2e, e2vid}, their integration into LVLMs remains limited. 
Recent event-LVLM methods such as
EventCLIP~\cite{eventclip}, EventBind~\cite{eventbind}, and
EventGPT~\cite{eventgpt} align event representations with vision-language
embeddings or generate event-based captions, but typically require dedicated
event encoders or additional training. In contrast, ECP uses events as a
training-free score-level motion and structure prior for pruning visual tokens
in existing RGB Video-LLMs, without parameter updates or event-embedding injection.

% \vspace{-0.4cm}
\section{Methodology}
\label{sec:methodology}

% Existing pruning methods~\cite{fastv, pyramiddrop, divprune, btp, dynamicvit} struggle to preserve both dynamics and geometry required for first-person dynamic spatial reasoning. Event cameras~\cite{gallego2020event, chakravarthi2024event_survey} provide an appealing alternative: their asynchronous sensing naturally captures motion and boundaries at negligible computational cost. The key challenge lies in bridging the representational gap between asynchronous event streams and the synchronous grid features of Video-LLMs~\cite{qwen2_5_vl, llava_onevision}. We address this with Event Cascade Pruning (ECP), a training-free framework illustrated in Figure~\ref{fig:framework}. ECP comprises three cascaded modules: Event-Triggered Causal Sampling (ETCS) for temporal keyframe selection, Event-guided Motion Saliency Filtering (EMSF) for spatial region filtering, and Event-Attention Ranking Fusion (EARF) for event-assisted pruning.

Existing pruning methods~\cite{fastv, pyramiddrop, divprune, btp, dynamicvit} struggle to preserve both dynamics and geometry required for first-person dynamic spatial reasoning. Event cameras~\citep{gallego2020event, chakravarthi2024event_survey} provide a
lightweight external cue by recording asynchronous brightness changes that reveal
motion activity and motion-induced structural cues. The key challenge lies in bridging the representational gap between asynchronous event streams and the synchronous grid features of Video-LLMs~\cite{qwen2_5_vl, llava_onevision}. We propose Event Cascade Pruning
(ECP), a training-free framework shown in Figure~\ref{fig:framework}, which
cascades temporal--spatial--semantic pruning: ETCS selects keyframes, EMSF
filters low-motion regions, and EARF fuses event saliency with attention-derived
semantic scores for layer-wise token pruning.

\subsection{Preliminaries}
\label{sec:preliminaries}
\textbf{Attention-based Pruning.}
Token pruning reduces computational overhead by retaining the top-$k$ tokens based on an importance score $\mathbf{s} \in \mathbb{R}^{T \times N}$. Existing methods~\cite{fastv,pyramiddrop,vtw} derive $\mathbf{s}$ from text-image attention weights:
\begin{equation}
\mathbf{S}_{\text{img}}^{(l)} = \frac{1}{m} \sum_{i=1}^{m} \text{Atten}^{(l)}(\mathbf{X}_I, \mathbf{X}_T^{(i)})
\end{equation}
where $\mathbf{X}_I$ and $\mathbf{X}_T$ denote image and text tokens, and $m$ is the number of text tokens.

\textbf{Event Camera as Motion Prior.}
Unlike frame-based sensors, event cameras~\cite{gallego2020event} asynchronously trigger events $e_k = (\mathbf{u}_k, t_k, p_k)$ only when logarithmic intensity changes exceed a threshold:
\begin{equation}
    |\ln I(\mathbf{u}, t) - \ln I(\mathbf{u}, t_{\text{prev}})| \geq \theta
\end{equation}
The resulting stream $\mathcal{E}$ offers microsecond-level temporal resolution, providing rich cues about motion causality and geometric boundaries that can be leveraged for spatial reasoning.

\subsection{Event-Triggered Causal Sampling}
\label{sec:etcs}

Existing sampling strategies struggle to balance coverage and efficiency: uniform sampling misses transient dynamics, while appearance-based methods incur high computational cost and are sensitive to motion blur. 
% We propose ETCS, a lightweight paradigm that leverages high-frequency event streams to anchor physical activity and causal transitions.
We propose ETCS, a lightweight paradigm that leverages high-temporal-resolution
event streams to anchor motion-rich moments and critical change points in scene
dynamics.

\textbf{Event Activity Flux.} To align temporal sampling with non-uniform motion activity, we divide the event stream into fixed temporal windows $W_n=[n\Delta t,(n+1)\Delta t)$ indexed by $n$. Applying a spatiotemporal density filter $\Phi(\cdot)$ to mitigate sensor noise, we define the window-level activity flux as $S_n$. A larger $S_n$ indicates stronger event activity within window $n$:
\begin{equation}
{
S_n =
\sum_{k=1}^{|\mathcal{E}|}
\Phi(e_k)\,
\mathbb{I}\!\left[t_k \in W_n\right].
}
\end{equation}

\textbf{Dual-Criteria Frame Anchoring.}
ETCS anchors frames using two complementary cues: high event activity $S_n$
preserves motion-rich moments, while large activity changes indicate critical
change points in scene dynamics:
\begin{equation}
\Delta S_n = |S_n-S_{n-1}|,\quad n\ge 1 .
\end{equation}
A large $\Delta S_n$ indicates that the event activity changes sharply
between adjacent windows, and is therefore treated as a critical temporal
anchor. Given the budget $N_{\text{target}}$, ETCS jointly selects high-$S_n$
windows and top-ranked $\Delta S_n$ windows, then refines the set by pruning
clustered low-activity candidates or filling large temporal gaps. The selected
windows are mapped to the nearest RGB keyframes. Since both cues are computed on
density-filtered window activity, ETCS suppresses event jitter while maintaining
temporal coverage.

\subsection{Event-guided Motion Saliency Filtering}
\label{sec:emsf}

Following temporal sampling, spatial redundancy remains a bottleneck. RGB-based
motion differencing requires additional frame processing and can degrade under
fast motion or blur, whereas event streams provide lightweight,
high-temporal-resolution cues of local brightness changes. EMSF distills the
event activity aligned with each selected keyframe into a token-aligned motion
prior, enabling early suppression of event-inactive visual tokens.

\textbf{Token-aligned Motion Saliency.}
For each selected keyframe $f$, let $\mathcal{W}_f=[a_f,a_f+\Delta t)$ denote
its aligned event window. Each visual token $i\in\{1,\ldots,N\}$ corresponds to
a spatial support $\Omega_i^{(f)}$ in the image plane. Given the density-filtered
event stream $\tilde{\mathcal E}$, we compute the event activity associated with
token $i$ as:
\begin{equation}
C_i^{(f)}
=
\sum_{e_k\in\tilde{\mathcal E}}
\mathbb{I}
\left[
(x_k,y_k)\in\Omega_i^{(f)},\
t_k\in\mathcal{W}_f
\right],
\quad
M_i^{(f)}=\mathcal{N}_f\!\left(C_i^{(f)}\right),
\end{equation}
where $\mathcal{N}_f(\cdot)$ denotes min-max normalization over all visual
tokens in keyframe $f$. The resulting score $M_i^{(f)}\in[0,1]$ measures the
event-guided motion saliency of visual token $i$.

\textbf{Budgeted Motion-token Retention.}
With retention ratio $\rho$, EMSF keeps $K=\lfloor\rho N\rfloor$ visual tokens
per keyframe by maximizing the retained motion saliency:
\begin{equation}
\mathcal{I}_{\mathrm{EMSF}}^{(f)}
=
\arg\max_{\mathcal{I}\subseteq\{1,\ldots,N\},\ |\mathcal{I}|=K}
\sum_{i\in\mathcal{I}} M_i^{(f)} .
\end{equation}
By reducing the number of visual tokens from $N$ to $K$ before subsequent
attention, EMSF reduces the visual-token attention term from $O(N^2)$ to
$O(K^2)=O(\rho^2N^2)$ and provides motion-salient candidates for EARF.

\subsection{Event-Attention Rank Alignment and Fusion}
\label{sec:earf}

% While EMSF filters textureless or low-dynamic regions, the remaining tokens require progressive sparsification to satisfy the stringent KV cache~\cite{xiao2024efficient, snapkv} and computational constraints of embodied agents. To ensure the integrity of the residual stream, we propose EARF, which calibrates semantic attention through an event-assisted pruning objective.

While EMSF suppresses event-inactive visual tokens, the remaining candidates
still dominate subsequent attention and KV-cache storage~\citep{xiao2024efficient,
snapkv}. EARF further sparsifies these tokens at selected pruning layers
$\mathcal{L}_{p}$ by calibrating attention-based importance with event saliency.
Importantly, events are used only to compute pruning scores and masks, rather
than being encoded, concatenated with RGB embeddings, or injected into hidden
representations.

\begin{wrapfigure}{r}{0.50\columnwidth}
    \centering
    \includegraphics[width=\linewidth]{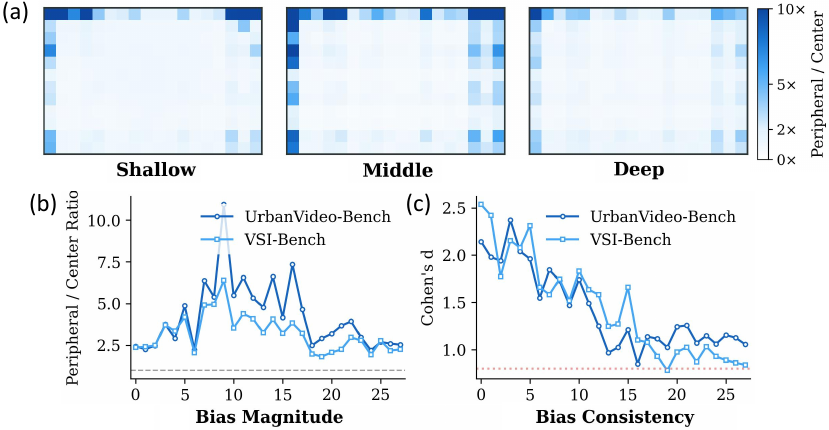}
    \vspace{-0.4cm}
    % \caption{\textbf{Peripheral Sink.} Visual attention is consistently biased
    % toward image borders across layers, with similar trends across datasets.}
    \caption{Peripheral Sink. (a) Attention maps reveal border-biased
    visual attention. (b) Peripheral-to-center ratio measures bias magnitude across
    layers. (c) Cohen's $d$ measures bias reliability, with similar trends across
    datasets.}    
    \vspace{-0.4cm}
    \label{fig:spatial_bias_combined}
\end{wrapfigure}

% \textbf{Empirical Motivation: The Peripheral Sink Pathology.}
% A quantitative analysis of 21,920 frames across 28 transformer layers 
% (Figure~\ref{fig:spatial_bias_combined}) reveals a systemic \textit{Peripheral Sink} 
% phenomenon in Video-LLMs~\cite{qwen2_5_vl}. This extends the attention sink in 
% LLMs~\cite{xiao2024efficient} to the spatial domain: attention concentrates toward 
% image borders rather than initial tokens, with peripheral-to-center ratios exceeding 
% unity across all layers ($p < 10^{-10}$). 
% The bias exhibits distinct layer-dependent patterns: shallow layers (0--7) show moderate 
% yet consistent bias (Cohen's $d = 1.97$, measuring effect size; CV $= 0.34$, coefficient of variation); middle layers (8--16) exhibit 
% the strongest bias ($5.64\times$) with high variance (CV $= 0.68$); deep layers (17--27) 
% display weaker bias ($2.86\times$) with content-dependent variation. 
% A strong cross-dataset correlation ($r = 0.897$) confirms this bias is model-intrinsic. 
% Pruning based on such biased scores retains positional artifacts at the expense of 
% task-relevant regions, motivating our event-assisted pruning calibration.

\textbf{Empirical Motivation: Peripheral Sink.}
Our analysis of $21{,}920$ frames across $28$ transformer layers
(Figure~\ref{fig:spatial_bias_combined}) reveals a consistent
\textit{Peripheral Sink} in Video-LLMs~\citep{qwen2_5_vl}: visual attention
concentrates more on image borders than on central regions, with the
peripheral-to-center attention ratio exceeding $1$ across all layers
($p<10^{-10}$). This spatial bias resembles the
attention sink observed in LLMs~\citep{xiao2024efficient}, but appears over
peripheral visual tokens rather than initial text tokens.  The bias is layer-dependent and is strongest in middle layers,
where the ratio reaches $5.64\times$. Moreover, the layer-wise bias profiles are
highly correlated across datasets ($r=0.897$), suggesting that the effect is
largely model-intrinsic rather than dataset-specific. As a result, pruning
solely by attention can preserve position-biased artifacts while discarding
motion-relevant regions, motivating our event-guided rank calibration. Detailed per-layer statistics are provided in Appendix \ref{appendix:attention_bias}.

\textbf{Fusion via Rank Projection.}
% Starting from the EMSF-retained candidate tokens, EARF further sparsifies visual tokens at selected pruning layers $\mathcal{L}_{p}$ to mitigate the peripheral bias identified above. For a pruning layer $l$ and frame $f$, let $\mathcal{V}^{(l,f)}$ denote the active visual-token set before pruning. EARF first extracts a semantic importance score from the preceding self-attention map:
Starting from the EMSF-retained candidates, EARF further sparsifies visual tokens
at pruning layers $\mathcal{L}_{p}$. For pruning layer $l$ and keyframe $f$, let
$\mathcal{V}^{(l,f)}$ denote the active visual-token set before pruning. 
EARF first derives a text-conditioned visual importance score from the preceding
post-softmax multimodal self-attention map:
% EARF first derives a text-conditioned visual importance score by aggregating
% query-to-visual attention from the preceding post-softmax multimodal
% self-attention map:
\begin{equation}
A_i^{(l,f)}
=
\frac{1}{|\mathcal{Q}_s|}
\sum_{q\in\mathcal{Q}_s}
\mathbf{A}^{(l-1)}[q,i],
\quad
i\in\mathcal{V}^{(l,f)} .
\end{equation}
% Here $\mathcal{Q}_s$ denotes the scoring query-token set used to read out visual
% importance, and $\mathbf{A}^{(l-1)}$ is the head-averaged attention map. The event-guided saliency $M_i^{(f)}$ is obtained from EMSF for the same visual
% token. Since attention scores and event saliency scores lie on different numerical
% scales and often exhibit long-tailed distributions (see Figure~\ref{fig:long_tail} in Appendix), direct arithmetic fusion can
% be dominated by a few extreme tokens. EARF therefore projects both signals into
% a shared rank space within each keyframe:
Here $\mathcal{Q}_s$ denotes the scoring query-token set used to read out visual
importance, and $\mathbf{A}^{(l-1)}$ is the head-averaged post-softmax attention map used for
scoring. The event-guided saliency $M_i^{(f)}$ is obtained from EMSF for the same
visual token. Since attention scores and event saliency scores lie on different
numerical scales and often exhibit long-tailed distributions (see Figure~\ref{fig:long_tail} in Appendix \ref{appendix:long_tail}), direct arithmetic
fusion can be dominated by a few extreme tokens. EARF therefore projects both
signals into a shared rank space within each keyframe:
\begin{equation}
R_{\mathcal{V}^{(l,f)}}(\phi,i)
=
\frac{
\operatorname{rank}_{\mathcal{V}^{(l,f)}}(\phi_i)
}{
\max(|\mathcal{V}^{(l,f)}|-1,1)
},
\quad
\phi\in\{A^{(l,f)},M^{(f)}\}.
\end{equation}
Here $\operatorname{rank}_{\mathcal{V}^{(l,f)}}(\phi_i)$ denotes the zero-based ascending rank of token $i$ among active visual tokens in $\mathcal{V}^{(l,f)}$, so larger scores receive larger normalized ranks. This non-parametric projection mitigates scale mismatch while preserving ordinal token importance.

\textbf{Layer-wise Rank Calibration and Pruning.}
EARF computes the calibrated pruning score as:
\begin{equation}
S_{\mathrm{calib}}^{(l,f,i)}
=
(1-\gamma_l)
R_{\mathcal{V}^{(l,f)}}(A^{(l,f)},i)
+
\gamma_l
R_{\mathcal{V}^{(l,f)}}(M^{(f)},i).
\end{equation}
% The weight $\gamma_l\in[0,1]$ controls the strength of event-based calibration.
% Rather than tuning an independent weight for every layer, we use a coarse
% layer-group schedule guided by the peripheral-bias profile in
% Figure~\ref{fig:spatial_bias_combined}: layers with stronger and more consistent
% peripheral bias receive larger event calibration, while deeper layers receive
% smaller calibration to preserve semantic flexibility.
% The weight $\gamma_l$ primarily follows the bias reliability measured by Cohen's $d$
% (Figure~\ref{fig:spatial_bias_combined}(c)): layers with more reliable
% peripheral-bias patterns receive stronger event-guided correction.

The weight $\gamma_l$ primarily follows the bias reliability measured by Cohen's $d$
(Figure~\ref{fig:spatial_bias_combined}(c)): high-$d$ layers
receive stronger event-guided correction, while deeper low-$d$ layers use
smaller $\gamma_l$ to preserve content-dependent semantic attention. 
% Middle stages retain moderate calibration because their bias magnitude remains large (Figure~\ref{fig:spatial_bias_combined}(b)).

Given the layer-wise retention ratio $\rho_l$, EARF keeps
$K_l^{(f)}=\max(1,\lfloor\rho_l|\mathcal{V}^{(l,f)}|\rfloor)$ visual tokens with
the highest calibrated scores in each keyframe:
\begin{equation}
\mathcal{I}_{\mathrm{EARF}}^{(l,f)}
=
\arg\max_{\mathcal{I}\subseteq\mathcal{V}^{(l,f)},\ |\mathcal{I}|=K_l^{(f)}}
\sum_{i\in\mathcal{I}} S_{\mathrm{calib}}^{(l,f,i)} .
\end{equation}
%  By down-weighting peripheral tokens, EARF increases the likelihood that they are pruned rather than retained, reducing the presence of positionally-biased keys and values in the KV cache. A theoretical analysis is provided in Appendix~\ref{sec:appendix_C}.
% Text and special tokens are always retained. Visual tokens outside $\mathcal{I}_{\mathrm{EARF}}^{(l,f)}$ are not forwarded to subsequent layers, so their keys and values no longer participate in later attention computation. 
By lowering the calibrated ranks of tokens
whose attention scores are inflated by peripheral bias but lack event support, EARF makes such tokens less likely to be retained, reducing the presence of
event-unsupported position-biased keys and values in the downstream KV cache. 
% A mechanistic analysis of the downstream KV-cache effect is provided in Appendix~\ref{sec:appendix_C}.
Appendix~\ref{sec:appendix_C} provides a mechanistic interpretation of how EARF reduces event-unsupported position-biased keys and values in subsequent attention computation.

\section{Experiments}
\label{sec:experiments}

We evaluate ECP on first-person dynamic spatial reasoning benchmarks using both simulated and real event streams. Our experiments address four questions: (1) Does ECP outperform existing pruning methods? (2) What are the computational gains? (3) Is ECP effective on both simulated and real events? (4) How does each component contribute?

% We evaluate ECP for first-person dynamic spatial reasoning in both simulated-event and native RGB-event settings, focusing on accuracy under different token budgets, inference efficiency, real-event generalization, and component ablations.

\subsection{Experimental Setup}
\label{sec:setup}

% \textbf{Benchmarks.}
% We focus on spatial reasoning during motion in 3D physical space. From two embodied video benchmarks---VSI-Bench~\cite{vsibench} (indoor navigation) and UrbanVideo-Bench~\cite{urbanvideobench} (outdoor drone scenarios)---we curate 13 challenging tasks that demand fine-grained geometric understanding: 9 from UrbanVideo-Bench (e.g., Landmark Position, Proximity) and 4 from VSI-Bench (e.g., Relative Distance, Route Planning). 
% Since these datasets lack event streams, we use vid2e~\cite{vid2e} to pre-generate them offline for benchmarking. 
% To further validate effectiveness on native event signals, we introduce ESR-Real, the first real-world RGB-event benchmark for first-person dynamic spatial reasoning, comprising over 700 QA pairs across 6 task categories (see Figure~\ref{fig:dataset_overview} for distribution). Dataset construction details are provided in Appendix~\ref{appendix:hardware}.

\textbf{Benchmarks.}
We focus on spatial reasoning during motion in 3D physical space and evaluate ECP in two event-stream settings.
For public RGB-only benchmarks, we use VSI-Bench~\cite{vsibench} (indoor navigation) and UrbanVideo-Bench~\cite{urbanvideobench} (outdoor drone scenarios). 
% We focus on 13 motion-centric tasks whose answers depend most directly on motion-conditioned geometry under ego-motion, spanning perception, cognition, and reasoning/planning (e.g., relative distance, proximity, route planning, and landmark positioning).  
We focus on 13 ego-motion-centric tasks covering perception, cognition, and reasoning/planning, such as relative distance, proximity, route planning, and landmark positioning.
Since these datasets lack native events, we generate simulated events offline with vid2e~\cite{vid2e} from each original video at its native frame rate. ETCS uses the full simulated event stream for temporal scoring, while EMSF/EARF only use event windows aligned with the retained RGB keyframes.
To further validate effectiveness on native event signals, we introduce ESR-Real, a synchronized RGB-event benchmark for first-person dynamic spatial reasoning with over 700 QA pairs across 6 categories (see Figure~\ref{fig:dataset_overview} for distribution). Dataset construction details are provided in Appendix~\ref{appendix:esr_real}.

\begin{wrapfigure}{r}{0.46\columnwidth}
    \centering
    \includegraphics[width=\linewidth]{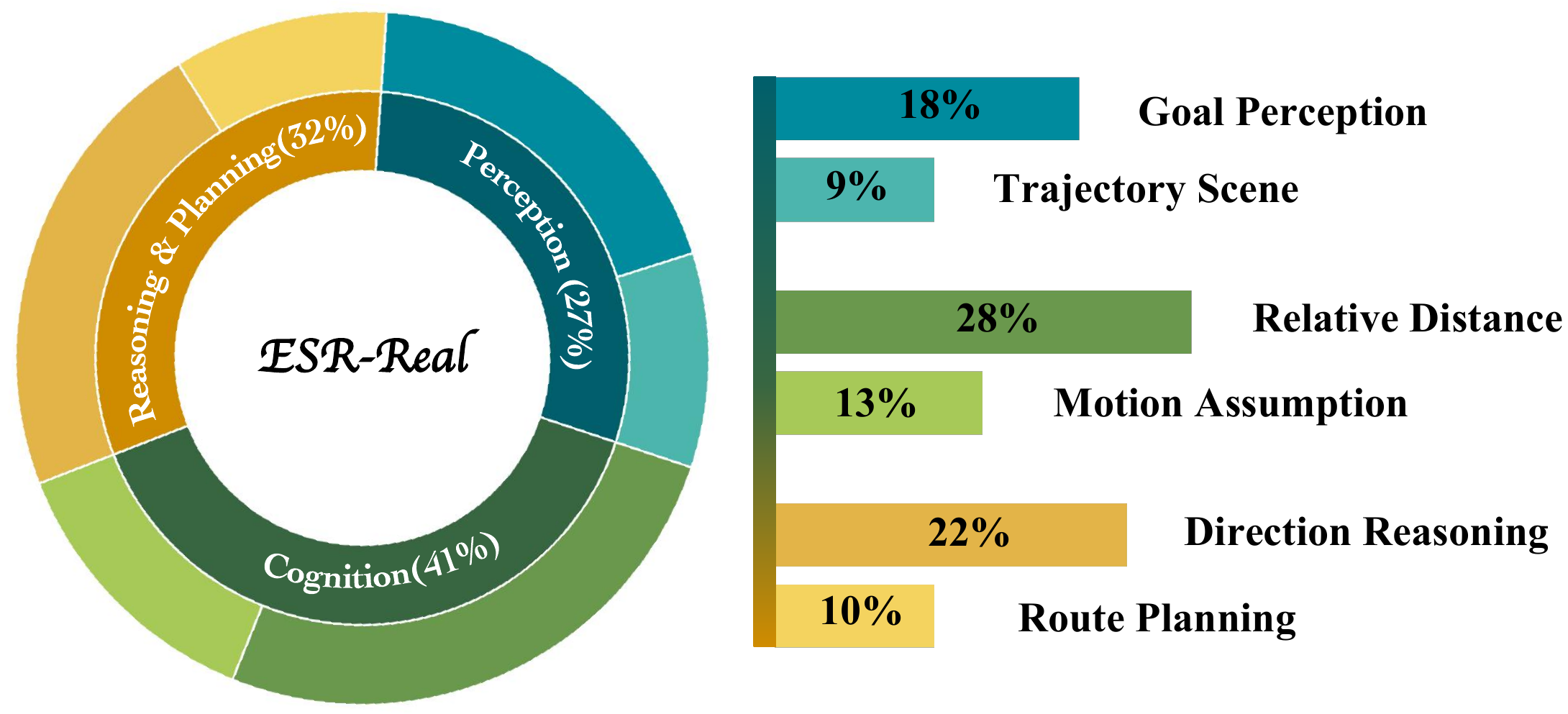}
        \vspace{-0.5cm}
    \caption{ESR-Real task distribution across Perception, Cognition, and Reasoning.}
        \vspace{-0.4cm}
    \label{fig:dataset_overview}
\end{wrapfigure}

\textbf{Baselines.}
We integrate ECP into Qwen2.5-VL-7B-Instruct~\cite{qwen2_5_vl} and compare with seven training-free baselines. 
Six of them are RGB-only pruning methods: DTD~\cite{dtd} uses inter-frame differencing; FastV~\cite{fastv}, PyramidDrop~\cite{pyramiddrop}, and VTW~\cite{vtw} use attention-based pruning; DivPrune~\cite{divprune} uses diversity-based pruning; and BTP~\cite{btp} combines attention saliency with diversity. 
We also include Direct Event Pruning (DEP), an event-only baseline that directly uses token-aligned event density maps as pruning saliency under the same token budgets. 
DEP controls for whether improvements come from event access alone. 
All methods are evaluated at \(70\%\), \(50\%\), and \(20\%\) final visual-token retention relative to the unpruned model.

\textbf{Implementation.}
Experiments run on NVIDIA A6000 GPUs with FlashAttention-2~\cite{flashattention2} acceleration. All methods are implemented on the same codebase for fair comparison.
% Following standard practice in VLM evaluation~\cite{qwen2_5_vl, llava_onevision, vsibench}, we use greedy decoding for deterministic inference, eliminating stochastic variation in model outputs.
Following standard practice in VLM evaluation~\cite{llava_onevision, vsibench, qwen2_5_vl}, we use greedy decoding for deterministic inference. 
% This ensures exact reproducibility across runs, as identical inputs yield identical outputs without random seed variation.
Event streams are aligned with visual tokens following the procedure described in Section~\ref{sec:emsf}. Pruning layers are determined by the layer-wise bias dynamics (Figure~\ref{fig:spatial_bias_combined}b): we intervene at stages preceding rapid bias escalation to preemptively correct attention drift before it propagates through the network. The fusion weight $\gamma_l$ is modulated according to bias consistency (Figure~\ref{fig:spatial_bias_combined}c): layers with high and stable Cohen's $d$ receive stronger event-guided motion prior correction, while layers exhibiting content-dependent variation preserve greater semantic flexibility. 
% In addition to accuracy, we report inference latency, GFLOPs, and peak memory.
We report inference latency, GFLOPs, and peak memory. For RGB-only public benchmarks, offline vid2e synthesis is excluded from the reported inference cost.

\definecolor{bg_perception}{RGB}{230, 242, 255} % 淡蓝色
\definecolor{bg_cognition}{RGB}{235, 250, 235}  % 淡绿色
\definecolor{bg_planning}{RGB}{255, 245, 230}   % 淡橙色
\definecolor{bg_ours}{RGB}{155,156,161}
\newcolumntype{C}{>{\centering\arraybackslash}p{2.3cm}}
\begin{table*}[t]
\caption{\textbf{Accuracy comparison on first-person dynamic spatial reasoning benchmarks}. Avg.Acc is computed over all 13 selected tasks; six representative task columns are shown here and the remaining seven are reported in Appendix~\ref{sec:remaining_results}. Best in bold. \textit{Task definitions}: \textbf{SE.Pos}=Start/End Position; \textbf{Goal.Det}=Goal Detection; \textbf{Rel.Dis}=Relative Distance; \textbf{Proximity}=Proximity Estimation; \textbf{Ass.Reason}=Association Reasoning; \textbf{Landmark.Pos}=Landmark Positioning.}
\label{tab:spatial_results_even}
\vspace{0.15cm}
\centering
% 消除颜色块上下的白边
\setlength{\aboverulesep}{0pt}
\setlength{\belowrulesep}{0pt}
% 适度减小行高，配合宽列实现“扁平”效果
\renewcommand{\arraystretch}{1.05}

\resizebox{\textwidth}{!}{
% 使用 l (左对齐) + 7个 C (固定等宽居中)
\begin{tabular}{l *{7}{C}}
\toprule

% --- 表头第一行 ---
\multirow{2}{*}{\textbf{Method}} & 
\multirow{2}{*}{\textbf{Avg.Acc}} & 
\multicolumn{2}{c}{\cellcolor{bg_perception}\textbf{Perception}} & 
\multicolumn{2}{c}{\cellcolor{bg_cognition}\textbf{Cognition}} & 
\multicolumn{2}{c}{\cellcolor{bg_planning}\textbf{Reasoning \& Planning}} \\

% --- 横向分隔线 ---
%\cmidrule{3-4} \cmidrule{5-6} \cmidrule{7-8}

% --- 表头第二行 ---
 & & SE.Pos & Goal.Det & Rel.Dis & Proximity & Ass.Reason & Landmark.Pos \\
\midrule

% --- Baseline ---
Original       & 36.31 & 55.10 & 39.79 & 44.51 & 50.00 & 16.90 & 37.41 \\
\midrule

% --- 70% Retaining ---
\multicolumn{8}{c}{\textit{70\% Retaining}} \\
\midrule
DTD            & 36.36 & 55.10 & 39.44 & 42.63 & 46.88 & 15.96 & 37.76 \\
FastV          & 36.28 & 57.14 & 39.08 & 44.51 & 50.00 & 16.90 & 37.41 \\
PDrop          & 36.73 & 53.06 & 40.49 & 44.83 & 46.88 & 15.96 & 36.71 \\
DivPrune       & 36.16 & 55.10 & 39.79 & 43.26 & 46.88 & 15.96 & 37.06 \\
BTP            & 35.95 & 57.14 & 39.44 & 43.26 & 46.88 & 17.37 & 36.71 \\
VTW            & 36.45 & 57.14 & 39.44 & 44.83 & 50.00 & 17.84 & 36.71 \\
DEP            & 36.88 & 53.10 & 38.30 & 45.80 & 50.00 & \textbf{18.80} & \textbf{37.80} \\
\rowcolor{gray!10}OURS          & \textbf{37.78} & \textbf{59.18} & \textbf{40.49} & \textbf{49.22} & \textbf{56.25} & 17.37 & 37.76 \\
\midrule

% --- 50% Retaining ---
\multicolumn{8}{c}{\textit{50\% Retaining}} \\
\midrule
DTD            & 36.55 & 59.18 & 39.44 & 42.01 & 53.12 & 17.37 & 35.31 \\
FastV          & 35.67 & 51.02 & 37.32 & 42.32 & 46.88 & 17.37 & 37.06 \\
PDrop          & 35.67 & 51.02 & 39.79 & 41.69 & 46.88 & 16.43 & \textbf{37.41} \\
DivPrune       & 36.77 & 48.98 & 39.79 & 44.51 & 46.88 & 18.31 & 35.31 \\
BTP            & 35.09 & 51.02 & 40.49 & 40.13 & 50.00 & 15.96 & 35.66 \\
VTW            & 33.84 & 42.86 & 33.45 & 38.87 & 53.12 & 14.08 & 34.97 \\
DEP            & 33.28 & 38.80 & 32.80 & 38.20 & 53.10 & 15.00 & 34.60 \\
\rowcolor{gray!10}OURS          & \textbf{38.16} & \textbf{59.18} & \textbf{42.25} & \textbf{50.47} & \textbf{53.12} & \textbf{18.78} & 37.06 \\
\midrule

% --- 20% Retaining ---
\multicolumn{8}{c}{\textit{20\% Retaining}} \\
\midrule
DTD            & 35.37 & 51.02 & 37.68 & 39.50 & 50.00 & 16.43 & 35.31 \\
FastV          & 32.64 & 28.57 & 29.58 & 33.23 & 46.88 & 14.55 & 31.82 \\
PDrop          & 33.12 & 26.53 & 31.69 & 33.86 & 46.88 & 15.96 & 32.52 \\
DivPrune       & 34.40 & 32.65 & 36.97 & 36.36 & 56.25 & 16.90 & 34.62 \\
BTP            & 33.34 & 30.61 & 33.10 & 35.11 & 56.25 & 12.68 & 33.92 \\
VTW            & 30.50 & 20.41 & 32.75 & 36.36 & 53.12 & 13.15 & 32.17 \\
DEP            & 29.95 & 22.40 & 30.70 & 36.10 & \textbf{59.40} & 12.20 & 30.10 \\
\rowcolor{gray!10}OURS          & \textbf{37.62} & \textbf{55.10} & \textbf{38.03} & \textbf{44.20} & 56.25 & \textbf{20.19} & \textbf{36.36} \\
\bottomrule
\end{tabular}
}

\end{table*}

\subsection{Main Results}
\label{sec:main_results}

\textbf{Accuracy Comparison.}
Table~\ref{tab:spatial_results_even} reveals distinct patterns across compression levels. At 70\% retention, all methods remain competitive---the large token pool provides redundancy for baselines to capture useful tokens. As compression intensifies, this tolerance vanishes: at 20\%, attention-based methods drop sharply (FastV: 32.64\%, VTW: 30.50\%), while ECP maintains 37.62\%, surpassing even the full-token baseline (36.31\%). Although DEP utilizes the same event modality,  it suffers severe accuracy degradation at 50\% (33.28\%) and 20\% (29.95\%) retention. At \(20\%\) retention, DEP trails ECP by \(7.67\) percentage points. 
This suggests that our performance gains stem from the proposed semantic alignment and cascade architecture, rather than an advantage from simply introducing the event modality.

Task-wise analysis reveals where event-assisted pruning excels. For motion-dependent tasks like Rel.Dis, ECP achieves 44.20\% at 20\% retention---nearly matching the unpruned baseline (44.51\%)---while baselines drop to 33--39\%. In geometry-reliant tasks like SE.Pos, DEP plummets to a mere 22.40\%. In contrast, ECP matches the full-token performance at \(55.10\%\). These tasks require continuous motion tracking and boundary perception, precisely what event cameras capture. In contrast, Landmark.Pos shows smaller margins, as landmark localization relies on static scene memory rather than dynamic cues.

\begin{table*}[h]
\caption{\textbf{Accuracy comparison on ESR-Real benchmark.} ECP demonstrates consistent effectiveness on real event streams captured by physical sensors. Best in bold. \textit{Task definitions}: \textbf{Goal.Percept}=Goal Perception; \textbf{Traj.Scene}=Trajectory Scene; \textbf{Rel.Dis}=Relative Distance; \textbf{Motion.Ass}=Motion Assumption; \textbf{Dir.Rea}=Direction Reasoning; \textbf{Route.Plan}=Route Planning.}
\label{tab:esr_real_results}
\centering
\setlength{\aboverulesep}{0pt}
\setlength{\belowrulesep}{0pt}
\renewcommand{\arraystretch}{1.05}
\resizebox{\textwidth}{!}{
\begin{tabular}{l *{7}{C}}
\toprule
\multirow{2}{*}{\textbf{Method}} & 
\multirow{2}{*}{\textbf{Avg.Acc}} & 
\multicolumn{2}{c}{\cellcolor{bg_perception}\textbf{Perception}} & 
\multicolumn{2}{c}{\cellcolor{bg_cognition}\textbf{Cognition}} & 
\multicolumn{2}{c}{\cellcolor{bg_planning}\textbf{Reasoning \& Planning}} \\
 & & Goal.Percept & Traj.Scene & Rel.Dis & Motion.Ass & Dir.Rea & Route.Plan \\
\midrule
Original       & 51.90 & 63.28 & 31.75 & 69.35 & 39.56 & 30.82 & 63.77 \\
\midrule
% --- 70% Retaining ---
\multicolumn{8}{c}{\textit{70\% Retaining}} \\
\midrule
DTD            & 51.20 & \textbf{65.62} & 28.57 & 67.34 & 35.16 & 30.82 & 66.67 \\
FastV          & 51.48 & 62.50 & 34.92 & 68.34 & 37.36 & 30.82 & 63.77 \\
PDrop          & 51.20 & 60.94 & 30.16 & 69.85 & 37.36 & 29.56 & 66.67 \\
DivPrune       & 50.92 & 62.50 & 33.33 & 68.84 & 38.46 & 27.67 & 63.77 \\
BTP            & 50.21 & 60.94 & 31.75 & 68.34 & 36.26 & 28.30 & 63.77 \\
VTW            & 50.78 & 57.81 & 31.75 & \textbf{69.85} & 37.36 & 29.56 & 66.67 \\
DEP            & 50.60 & 60.20 & 28.60 & 68.30 & 37.40 & 29.60 & 68.10 \\
\rowcolor{gray!10}\textbf{OURS} & \textbf{52.89} & 64.06 & \textbf{38.10} & 68.84 & \textbf{39.56} & \textbf{30.82} & \textbf{68.12} \\
\midrule

% --- 50% Retaining ---
\multicolumn{8}{c}{\textit{50\% Retaining}} \\
\midrule
DTD            & 49.51 & 64.06 & 28.57 & 65.83 & 34.07 & 26.42 & 68.12 \\
FastV          & 48.94 & 63.28 & 31.75 & 64.32 & 36.26 & 26.42 & 62.32 \\
PDrop          & 50.49 & 63.28 & 33.33 & 65.83 & 37.36 & 29.56 & 63.77 \\
DivPrune       & 50.21 & 65.62 & 30.16 & 64.82 & 40.66 & 27.67 & 62.32 \\
BTP            & 50.21 & 63.28 & 34.92 & 64.32 & 38.46 & 28.30 & 65.22 \\
VTW            & 43.58 & 64.84 & 26.98 & 47.24 & 31.87 & 28.30 & 59.42 \\
DEP            & 41.50 & 64.80 & 27.00 & 42.70 & 27.50 & 27.00 & 59.40 \\
\rowcolor{gray!10}\textbf{OURS} & \textbf{54.58} & \textbf{67.97} & \textbf{38.10} & \textbf{69.85} & \textbf{40.66} & \textbf{32.70} & \textbf{69.57} \\
\midrule

% --- 20% Retaining ---
\multicolumn{8}{c}{\textit{20\% Retaining}} \\
\midrule
DTD            & 46.40 & 66.41 & 25.40 & 55.78 & 34.07 & 27.04 & 62.32 \\
FastV          & 39.77 & 66.41 & 20.63 & 37.19 & 34.07 & 25.79 & 55.07 \\
PDrop          & 39.49 & 66.41 & 23.81 & 29.65 & 38.46 & 28.30 & 59.42 \\
DivPrune       & 45.56 & 67.19 & 28.57 & 48.24 & 35.16 & 31.45 & 59.42 \\
BTP            & 43.30 & 66.41 & 20.63 & 49.25 & 34.07 & 25.16 & 57.97 \\
VTW            & 39.07 & 62.50 & 23.81 & 28.64 & 35.16 & 33.96 & 56.52 \\
DEP            & 40.20 & 65.60 & 28.60 & 29.10 & 37.40 & \textbf{34.60} & 52.20 \\
\rowcolor{gray!10}\textbf{OURS} & \textbf{50.07} & \textbf{67.19} & \textbf{31.75} & \textbf{60.30} & \textbf{38.46} & 28.93 & \textbf{69.57} \\
\bottomrule
\end{tabular}
}
\end{table*}

% To validate effectiveness on real sensors, we evaluate on ESR-Real (Table~\ref{tab:esr_real_results}). ECP's advantage amplifies: at 50\% retention, ECP achieves 54.58\% versus baseline 51.90\%; at 20\%, ECP (50.07\%) outperforms the second-best (DTD: 46.40\%) by 3.67\%.

To validate effectiveness on real sensors, we evaluate on ESR-Real (Table~\ref{tab:esr_real_results}). At 50\% retention, ECP achieves 54.58\%, surpassing the full-token baseline (51.90\%) by 2.68\%. Under more aggressive 20\% retention, ECP remains the most robust pruning method, outperforming the second-best baseline (DTD: 46.40\%) by 3.67\%.

\noindent
\begin{minipage}[t]{0.525\columnwidth}
% \textbf{Efficiency Analysis.}
% Table~\ref{tab:urban_efficiency_single} reports inference latency, GFLOPs, and memory usage. At 20\% retention, ECP achieves 1.11s latency (1.89$\times$ speedup), 141.9 GFLOPs (52\% reduction), and 16637 MB memory. Unlike DTD's pixel-wise differencing or DivPrune's $O(N^2)$ pairwise distances, ECP's sparse event accumulation provides motion-aware guidance with negligible overhead.
%  \vspace{-0.4cm}

\textbf{Efficiency Analysis.}
Table~\ref{tab:urban_efficiency_single} reports online inference cost with event streams available either natively or precomputed.  At 20\% retention, ECP achieves 1.11s latency (\(1.89\times\) speedup), 141.9 GFLOPs (52\% reduction), and 16637 MB memory. Unlike DTD’s pixel-wise differencing or DivPrune’s $O(N^2)$ pairwise distances, ECP’s sparse event accumulation provides motion-aware guidance with low overhead. For RGB-only public benchmarks, offline vid2e generation is used only to construct the simulated-event evaluation setting and is excluded from the reported inference cost.

\end{minipage}\hfill
\begin{minipage}[t]{0.49\columnwidth}
\centering
%  \vspace{-0.4cm}
\sidetablecaption{tab:urban_efficiency_single}{Efficiency comparison on the UrbanVideo-Bench (20\% retention ratio).}
\scriptsize
\setlength{\tabcolsep}{4pt}
{
\begin{tabular}{@{}lcccc@{}}
\toprule
Method & Ratio & Time (s) & GFLOPs $\downarrow$ & Mem (MB) \\
\midrule
Full Tokens & - & 2.10 & 294.6 & 18848 \\
\midrule
DTD & 20\% & 1.80 & 185.9 & 16880 \\
FastV & 20\% & 1.16 & 173.9 & 17068 \\
VTW & 20\% & 1.18 & 189.8 & 17200 \\
PDrop & 20\% & 1.24 & 187.5 & 17169 \\
DivPrune & 20\% & 1.25 & 187.1 & 17169 \\
BTP & 20\% & 1.55 & 187.5 & 17169 \\
\midrule
\rowcolor{gray!10} \textbf{Ours} & 20\% & \textbf{1.11} & \textbf{141.9} & \textbf{16637} \\
\bottomrule
\end{tabular}
}
\end{minipage}

\medskip

% \subsubsection{System Efficiency Analysis}
% As shown in Table~\ref{tab:ablation_study} (right columns), ECP significantly reduces computational overhead.
% \textbf{Latency Reduction:} Compared to the Baseline (1.96s), ECP reduces inference latency to \textbf{1.35s} (-31\%), making it feasible for real-time embodied control.
% \textbf{Zero-Cost Overhead:} A key concern with multi-modal fusion is the preprocessing cost. Our results show that event processing (ETCS + EMSF) adds negligible latency (<5ms), as it operates on sparse matrices without heavy neural encoders. This highlights the "Plugin-and-Play" nature of ECP.

\begin{figure*}[h]
  \centering
   \vspace{-0.25cm}
  % 0.9\linewidth 表示占页面总宽度的 90%
  \includegraphics[width=0.99\linewidth]{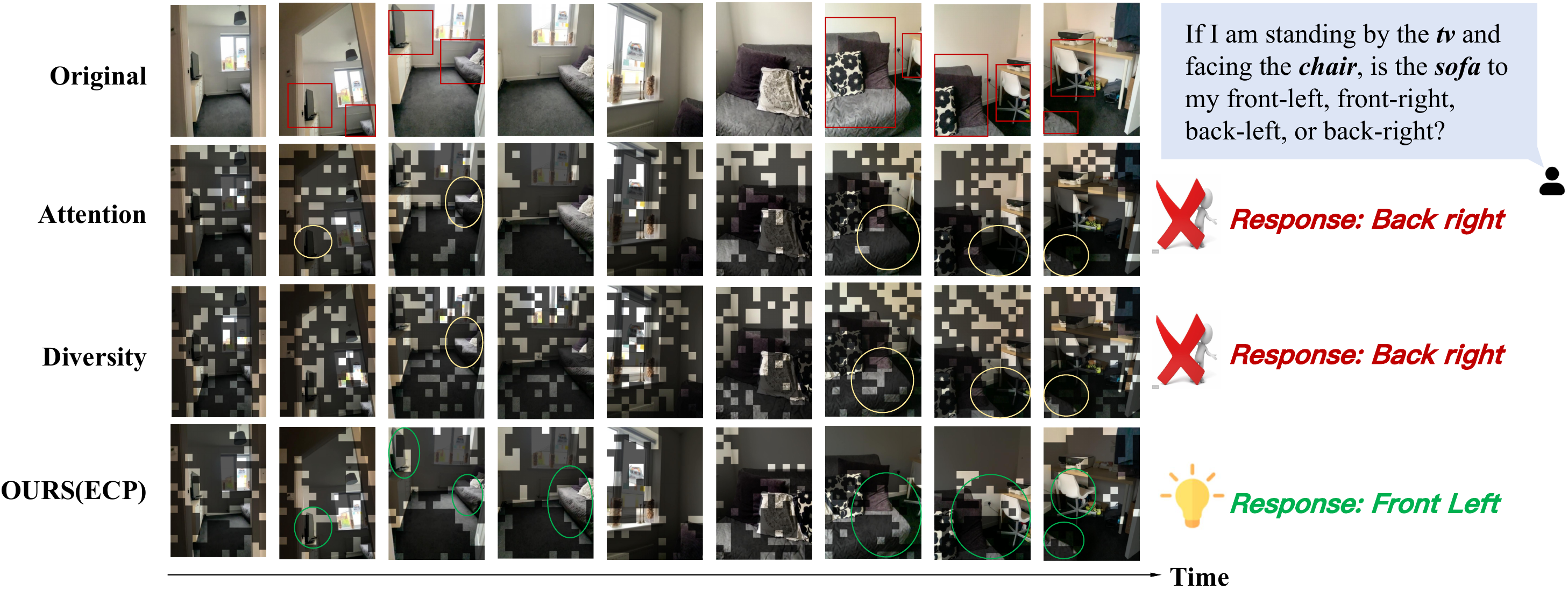}
  \vspace{-0.2cm} % 稍微收缩图片与标题之间的垂直距离，可根据需要调整
  \caption{\textbf{Qualitative comparison.} ECP retains physical structures even at high token compression.}
  \label{fig:qualitative_comparison}
  \vspace{-0.1cm} % 稍微收缩标题与正文之间的垂直距离
\end{figure*}

\textbf{Qualitative Analysis.}
Figure~\ref{fig:qualitative_comparison} visualizes token retention under high compression. Attention-based pruning (row 2) concentrates tokens at image borders, discarding the central sofa and chair critical for spatial judgment. Diversity-based pruning (row 3) fragments object boundaries into isolated patches, breaking geometric continuity. ECP (row 4) anchors tokens at regions exhibiting high motion saliency, preserving continuous object contours and enabling correct spatial reasoning.

% \subsection{\new{ESR-Real Benchmark Results}}
% \old{\subsection{ESR-Real Dataset}}

% \old{The experiments above rely on vid2e-simulated events. A natural question follows: does ECP generalize to native event signals with authentic noise characteristics? To investigate this, we introduce ESR-Real, a real-world benchmark captured with synchronized RGB-event cameras, comprising over 700 QA pairs across 6 task categories. Unlike simulated streams, ESR-Real contains native events with sensor-specific noise patterns and temporal jitter inherent to physical hardware. Details on hardware setup and calibration are provided in Appendix~\ref{appendix:hardware}. As shown in Figure~\ref{fig:dataset_overview}, ESR-Real covers diverse spatial reasoning tasks spanning Perception (27\%), Cognition (41\%), and Reasoning (32\%) categories.}

% \old{Table~\ref{tab:esr_real_results} presents the results. ECP not only maintains but amplifies its advantage on real-world data: at 50\% retention, ECP achieves 54.58\%, surpassing the full-token baseline (51.90\%) by 2.68\% while all other methods fall below baseline. At 20\% retention, ECP (50.07\%) outperforms the second-best (DTD, 46.40\%) by 3.67\%. This confirms that our physics-driven design is robust to real sensor noise and transfers effectively from simulation to reality.}

\subsection{Ablation Studies}
\label{sec:ablation_studies}

We conduct ablation studies to validate each component and design choice, examining three aspects: module contribution, fusion weight strategy, and layer selection.
% We ablate three factors: module composition, EARF fusion weights, and layer selection.

\noindent
\begin{minipage}[t]{0.665\columnwidth}
% \textbf{Component Contribution.} Table~\ref{tab:component_ablation_study} isolates the effect of each module. Due to space constraints, we provide a more granular component-wise analysis across specific task categories (Perception, Cognition, and Reasoning) in Appendix~\ref{sec:appendix_ablation}. EMSF alone yields a +0.91\% accuracy gain by filtering low-dynamic regions, confirming that event-based spatial filtering preserves physically salient tokens. EARF alone shows limited improvement, yet combining it with ETCS produces a +1.10\% gain. This synergy suggests that attention calibration requires temporally coherent keyframes to be effective---random frames lack the physical continuity needed for meaningful fusion. The full ECP achieves the best accuracy (37.85\%) with lowest latency (1.16s), demonstrating that the three modules form a complementary cascade.
\textbf{Component Contribution.} Table~\ref{tab:component_ablation_study} reports component ablations averaged over the three retention ratios. ETCS mainly improves efficiency, reducing latency from \(2.08\)s to \(1.58\)s while slightly improving accuracy by \(0.21\) points. EMSF and EARF improve accuracy by \(0.91\) and \(0.84\) points individually. Adding ETCS to EMSF or EARF further improves accuracy, while EMSF+EARF without ETCS is less effective, suggesting that temporal anchoring is important for coherent event-attention fusion. The full cascade achieves the best measured trade-off, with \(37.85\%\) accuracy and \(1.16\)s latency. Appendix~\ref{sec:appendix_ablation} provides task-category breakdowns.
\end{minipage}\hfill
\begin{minipage}[t]{0.305\columnwidth}
\centering
\vspace{-0.2cm}
\sidetablecaption{tab:component_ablation_study}{Component ablation.}
\scriptsize
\setlength{\tabcolsep}{3pt}
\renewcommand{\arraystretch}{1.03}
{
\begin{tabular}{@{} c c c c c@{}}
\toprule
 \textbf{ETCS} & \textbf{EMSF} & \textbf{EARF} & \textbf{Lat.(s)} & \textbf{Acc.(\%)} \\
\midrule
 - & - & - & 2.08 & 36.31 \\
\midrule
\checkmark & - & - & 1.58 & 36.52 \\
 - & \checkmark & - & 1.83 & 37.22 \\
 - & - & \checkmark & 1.53 & 37.15 \\
 \checkmark & \checkmark & - & 1.47 & 37.51 \\
\checkmark & - & \checkmark & 1.17 & 37.75 \\
 - & \checkmark & \checkmark & 1.59 & 37.00 \\
\midrule
\rowcolor{gray!10}  \checkmark & \checkmark & \checkmark & \textbf{1.16} & \best{37.85} \\
\bottomrule
\end{tabular}
}

\end{minipage}

\medskip

\noindent
\begin{minipage}[t]{0.539\columnwidth}
% \textbf{Fusion Weight Analysis.} Table~\ref{tab:systematic_ablation_final} examines the layer-wise fusion weight $\gamma_l$. Three findings stand out: (1) Increasing event prior weight consistently improves accuracy (34.18\% $\rightarrow$ 37.62\%), validating the importance of event-guided motion priors. (2) Pure event guidance ($\gamma$=1.0) slightly underperforms balanced fusion ($\gamma$=0.7), indicating semantic attention remains complementary. (3) The Decay pattern ($\gamma$: 0.8$\rightarrow$0.5$\rightarrow$0.3) outperforms Growth (0.3$\rightarrow$0.5$\rightarrow$0.8). This aligns with our bias analysis (Figure~\ref{fig:spatial_bias_combined}c): shallow layers exhibit high Cohen's $d$, indicating consistent and reliable bias that permits aggressive event-based correction; deep layers show lower Cohen's $d$ with content-dependent variation, requiring reduced $\gamma$ to avoid over-correction.
\textbf{Fusion Weight Analysis.} Table~\ref{tab:systematic_ablation_final} examines the layer-wise fusion weight $\gamma_l$ with pruning layers fixed. Three trends emerge: (1) Increasing the uniform event weight improves accuracy from 34.18\% to 37.47\% up to $\gamma=0.7$, while event-only rank fusion ($\gamma=1.0$) slightly drops to 37.36\%, indicating that semantic attention remains complementary. (2) The shallow-heavy Decay pattern ($0.8\rightarrow0.5\rightarrow0.3$) outperforms Growth ($0.3\rightarrow0.5\rightarrow0.8$) by 1.05 points. (3) The bias-guided schedule ($0.8,0.6,0.5$) achieves the best accuracy of 37.62\%. This aligns with Figure~\ref{fig:spatial_bias_combined}(c): high-$d$ shallow layers receive stronger event-guided calibration, whereas lower-$d$ deeper layers use smaller $\gamma_l$ to preserve content-dependent semantic attention.
\end{minipage}\hfill
\begin{minipage}[t]{0.43\columnwidth}
\centering
\vspace{-0.25cm}
\sidetablecaption{tab:systematic_ablation_final}{Layer-wise fusion-weight ablation at 20\% retention (S/M/D: shallow/mid/deep).}
\scriptsize
\setlength{\tabcolsep}{2pt}
\renewcommand{\arraystretch}{1.08}
{
\begin{tabular}{@{}c c c c c c@{}}
\toprule
\textbf{Strategy} & $\gamma_{\mathrm{S}}$ & $\gamma_{\mathrm{M}}$ & $\gamma_{\mathrm{D}}$ & \textbf{Acc.(\%)} & \textbf{Analysis} \\
\midrule
\multirow{5}{*}{Uniform} 
 & 0.0 & 0.0 & 0.0 & 34.18 & Baseline (Attn) \\
 & 0.3 & 0.3 & 0.3 & 35.66 & Weak Fusion \\
 & 0.5 & 0.5 & 0.5 & 36.21 & Medium Fusion \\
 & \textbf{0.7} & \textbf{0.7} & \textbf{0.7} & \textbf{37.47} & \textbf{Strong Fusion} \\
 & 1.0 & 1.0 & 1.0 & 37.36 & Pure Event \\
\midrule
Peak ($\wedge$) & 0.2 & 0.6 & 0.2 & 34.60 & Mid-focus \\
Valley ($\vee$) & \textbf{0.6} & 0.2 & 0.6 & 36.32 & Ends-focus \\
\midrule
Growth ($\nearrow$) & 0.3 & 0.5 & 0.8 & 35.99 & Suboptimal \\
Decay ($\searrow$) & \textbf{0.8} & 0.5 & 0.3 & 37.04 & \textbf{Supports Hyp.} \\
\midrule
\rowcolor{gray!10} \textbf{Best} & \textbf{0.8} & \textbf{0.6} & \textbf{0.5} & \textbf{37.62} & \textbf{Best Balance} \\
\bottomrule
\end{tabular}
}
\end{minipage}

\medskip

\noindent
\begin{minipage}[t]{0.67\columnwidth}

\textbf{Layer Selection Strategy.}
Table~\ref{tab:layer_selection_ablation_study} compares our bias-guided layer selection against uniform layer spacing, both using the same ECP modules and final token budgets. Uniform is marginally higher at 70\% retention (38.10 vs.\ 37.78),
suggesting a token-buffer regime where near-full spatial coverage can
preserve task-critical cues and mask suboptimal layer placement. Once
this buffer vanishes, stability becomes decisive: uniform spacing drops
3.24 points from 70\% to 20\% retention (38.10$\to$34.86), whereas our
bias-guided placement drops only 0.16 points (37.78$\to$37.62). These suggest that bias-guided layer selection mainly targets extreme-sparsity stability by
intervening before peripheral-bias escalation propagates through the
residual stream and KV cache.

\end{minipage}\hfill
\begin{minipage}[t]{0.4\columnwidth}
\centering
\sidetablecaption{tab:layer_selection_ablation_study}{Layer selection ablation.}
\scriptsize
\setlength{\tabcolsep}{4pt}
\renewcommand{\arraystretch}{1.05}
{
\begin{tabular}{@{}l l c c c@{}}
\toprule
\textbf{Ratio} & \textbf{Variant} & \textbf{Avg.} & \textbf{Urban} & \textbf{VSI} \\
\midrule
Full & Baseline & 36.31 & 40.86 & 32.03 \\
\midrule
70\% & Uniform & \textbf{38.10} & \textbf{42.91} & 33.13 \\
\rowcolor{gray!10} 70\% & \textbf{Ours} & 37.78 & 42.49 & \best{33.35} \\
\midrule
50\% & Uniform & 37.09 & 42.56 & 29.88 \\
\rowcolor{gray!10} 50\% & \textbf{Ours} & \best{38.16} & \best{42.85} & \best{33.76} \\
\midrule
20\% & Uniform & 34.86 & 39.99 & 29.77 \\
\rowcolor{gray!10} 20\% & \textbf{Ours} & \best{37.62} & \best{42.78} & \best{32.76} \\
\bottomrule
\end{tabular}
}
\end{minipage}

\medskip

% \section{Conclusion}
% We presented Event Cascade Pruning (ECP), the first training-free event-driven pruning framework for efficient first-person dynamic spatial reasoning. Existing pruning paradigms fail to preserve the dynamics and geometry essential for spatial reasoning: inter-frame methods miss continuous motion while intra-frame methods destroy geometric structures. ECP addresses this by deeply integrating event cameras into a cascaded temporal-to-spatial filtering pipeline. Through progressive keyframe anchoring, low-dynamic region filtering, and physics-semantic rank fusion, ECP translates the high-frequency event-guided motion priors into principled pruning guidance. 
% Experiments on first-person dynamic spatial reasoning benchmarks demonstrate that ECP achieves 80\% token reduction while surpassing the full-token baseline (37.62\% vs. 36.31\%), with 1.89$\times$ speedup and 52\% GFLOPs reduction. 
% Crucially, results on our proposed ESR-Real benchmark further confirm ECP's effectiveness on native event streams and its robustness under high compression.
% To facilitate reproducibility, we will release our code and the ESR-Real dataset upon publication.

\section{Conclusion and Limitations}
\label{sec:conclusion_limitations}
% We presented Event Cascade Pruning (ECP), a training-free
% event-assisted pruning framework for efficient first-person dynamic
% spatial reasoning. By cascading event-triggered keyframe anchoring,
% motion-saliency filtering, and event-attention rank fusion, ECP converts
% high-frequency event cues into pruning decisions that preserve dynamic
% and geometric structures under high compression. Experiments on public
% benchmarks and ESR-Real show that ECP maintains strong accuracy while reducing
% latency, GFLOPs, and memory, and we will release the code and ESR-Real
% dataset upon publication.

We presented Event Cascade Pruning (ECP), a training-free
event-assisted pruning framework for efficient first-person dynamic
spatial reasoning. By cascading event-triggered sampling,
motion-saliency filtering, and event-attention rank fusion, ECP converts
high-frequency event cues into pruning decisions that preserve dynamic
and geometric structures under high compression. At \(20\%\) token
retention, ECP surpasses the full-token baseline (\(37.62\%\) vs.\
\(36.31\%\)) with \(1.89\times\) speedup and \(52\%\) GFLOPs reduction;
on ESR-Real, it gains \(2.68\) points at \(50\%\) retention. We will release the code and ESR-Real
dataset upon publication.

\paragraph{Limitations.}
ECP assumes synchronized RGB-event inputs; it targets
platforms where an event camera is paired with RGB sensing, rather than
serving as an RGB-only online acceleration method. Although event sensors remain less ubiquitous than RGB cameras, commercialization
by Sony and Prophesee makes this setting increasingly feasible. More broadly, ECP illustrates sensor-assisted pruning and motivates future
external-prior-guided or RGB-only distilled pruning schemes.

{
\small
\bibliographystyle{plainnat}
\bibliography{ref}
}

%%%%%%%%%%%%%%%%%%%%%%%%%%%%%%%%%%%%%%%%%%%%%%%%%%%%%%%%%%%%

\appendix

\newpage
%%%%%%%%%%%%%%%%%%%%%%%%%%%%%%%%%%%%%%%%%%%%%%%%%%%%%%%%%%%%%%%%%%%%%%%%%%%%%%
%%%%%%%%%%%%%%%%%%%%%%%%%%%%%%%%%%%%%%%%%%%%%%%%%%%%%%%%%%%%%%%%%%%%%%%%%%%%%%%
%%%%%%%%%%%%%%%%%%%%%%%%%%%%%%%%%%%%%%%%%%%%%%%%%%%%%%%%%%%%%%%%%%%%%%%%%%%%%%%
% Appendix A: Spatial Attention Bias Analysis
%%%%%%%%%%%%%%%%%%%%%%%%%%%%%%%%%%%%%%%%%%%%%%%%%%%%%%%%%%%%%%%%%%%%%%%%%%%%%%%
\section{Spatial Attention Bias Analysis}
\label{appendix:attention_bias}

This appendix provides statistical analysis of the spatial attention bias in vision-language models. We analyze 21,920 frames across all 28 transformer layers from UrbanVideo-Bench (472 videos, 15,104 frames) and VSI-Bench (213 videos, 6,816 frames).

\subsection{Spatial Region Definition}
\label{appendix:region_def}

\begin{figure}[ht]
    \centering
    \includegraphics[width=0.45\linewidth]{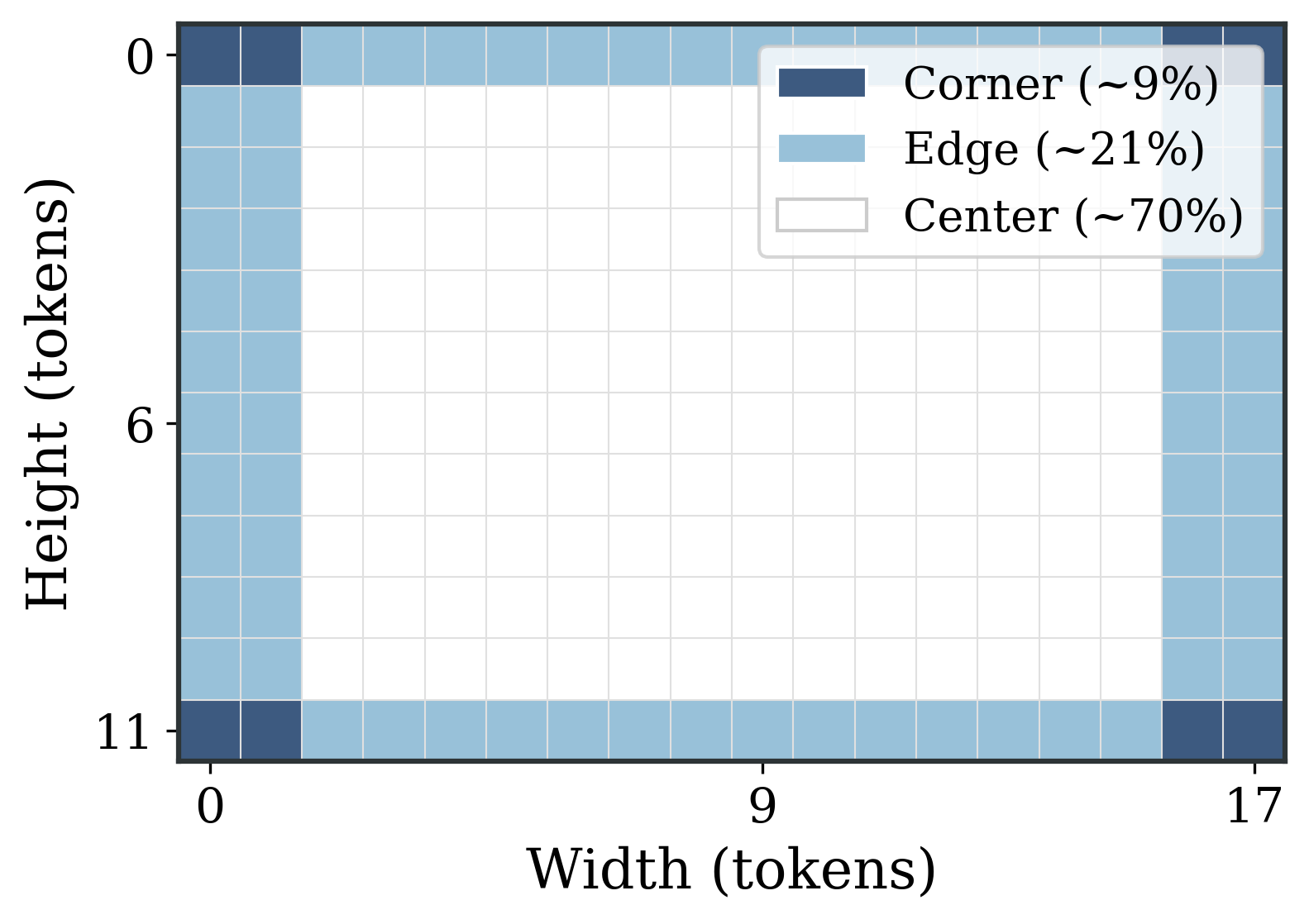}
    \caption{Spatial region partitioning. Peripheral = Corner (3.7\%) + Edge (31.5\%); Center (64.8\%) serves as baseline.}
    \label{fig:region_definition}
\end{figure}

All attention patterns are standardized onto a $12 \times 18$ spatial grid (216 tokens). As illustrated in Figure~\ref{fig:region_definition}, we partition this grid into three regions using a 15\% margin threshold: Corner (8 tokens, 3.7\%), Edge (68 tokens, 31.5\%), and Center (140 tokens, 64.8\%). The Peripheral region (Corner $\cup$ Edge, 76 tokens, 35.2\%) serves as the primary metric for quantifying spatial attention bias.

\subsection{Statistical Analysis}
\label{appendix:stats}

Table~\ref{tab:all_layer_stats} provides complete per-layer statistics; Table~\ref{tab:layer_group_stats} summarizes bias characteristics by layer group.

\begin{table*}[ht]
\centering
\caption{Complete layer-wise statistics ($n = 21{,}920$). $t$: one-sample $t$-statistic against $\mu_0 = 1.0$; all layers satisfy $p<10^{-10}$.}
\label{tab:all_layer_stats}
\begin{small}
\begin{tabular}{lccccc|lccccc}
\toprule
\textbf{L} & \textbf{Corner} & \textbf{Edge} & \textbf{Periph.} & \textbf{$d$} & \textbf{$t$} & \textbf{L} & \textbf{Corner} & \textbf{Edge} & \textbf{Periph.} & \textbf{$d$} & \textbf{$t$} \\
\midrule
0 & 7.70 & 1.80 & 2.41 & 2.24 & 331 & 14 & 13.70 & 4.90 & 5.82 & 0.99 & 146 \\
1 & 7.00 & 1.80 & 2.31 & 2.09 & 310 & 15 & 9.00 & 3.30 & 3.87 & 1.23 & 182 \\
2 & 7.60 & 1.90 & 2.48 & 1.88 & 279 & 16 & 12.60 & 5.50 & 6.25 & 0.80 & 118 \\
3 & 12.30 & 2.70 & 3.75 & \textbf{2.30} & 341 & 17 & 9.70 & 3.50 & 4.20 & 1.08 & 160 \\
4 & 9.20 & 2.30 & 3.05 & 2.01 & 297 & 18 & 4.00 & 2.10 & 2.33 & 1.04 & 154 \\
5 & 16.80 & 3.20 & 4.66 & 1.99 & 295 & 19 & 3.80 & 2.40 & 2.57 & 0.91 & 134 \\
6 & 6.40 & 1.70 & 2.18 & 1.56 & 231 & 20 & 5.00 & 2.60 & 2.84 & 1.10 & 163 \\
7 & 20.30 & 4.20 & 5.92 & 1.72 & 254 & 21 & 8.10 & 2.70 & 3.23 & 1.11 & 165 \\
8 & 19.40 & 3.60 & 5.26 & 1.73 & 256 & 22 & 10.20 & 2.90 & 3.64 & 1.00 & 148 \\
9 & \textbf{47.70} & 5.00 & \textbf{9.53} & 1.35 & 200 & 23 & 7.30 & 2.40 & 2.92 & 1.11 & 164 \\
10 & 18.90 & 3.20 & 4.89 & 1.59 & 235 & 24 & 4.40 & 1.80 & 2.11 & 1.01 & 150 \\
11 & 18.00 & 4.50 & 5.89 & 1.41 & 209 & 25 & 6.30 & 2.30 & 2.72 & 1.04 & 154 \\
12 & 12.40 & 4.10 & 4.95 & 1.26 & 187 & 26 & 5.70 & 2.10 & 2.47 & 1.04 & 153 \\
13 & 8.00 & 3.90 & 4.31 & 0.95 & 141 & 27 & 5.90 & 2.00 & 2.44 & 0.98 & 145 \\
\bottomrule
\end{tabular}
\end{small}
\end{table*}

\begin{table}[ht]
\centering
\caption{Summary statistics by layer group.}
\label{tab:layer_group_stats}
\begin{small}
\begin{tabular}{lccccc}
\toprule
\textbf{Group} & \textbf{Layers} & \textbf{Periph.} & \textbf{Corner} & \textbf{Edge} & \textbf{$d$} \\
\midrule
Shallow & 0--7 & $3.35\times$ & $10.20\times$ & $2.40\times$ & 1.97 \\
Middle & 8--16 & $5.64\times$ & $22.80\times$ & $4.20\times$ & 1.26 \\
Deep & 17--27 & $2.86\times$ & $6.10\times$ & $2.30\times$ & 1.04 \\
\bottomrule
\end{tabular}
\end{small}
\end{table}

Key findings from Table~\ref{tab:all_layer_stats}: (1) All 28 layers show significant peripheral bias ($p < 10^{-10}$). (2) Layer 9 exhibits peak ratio ($9.53\times$); Layer 3 achieves highest $d$ (2.30). (3) Mean ratio across layers is $3.89\times$. (4) Cross-dataset correlation $r = 0.897$ ($p < 10^{-10}$) confirms this is a model-intrinsic property.

\subsection{Effect Size Interpretation}
\label{appendix:effect_size}

\begin{figure}[H]
    \centering
    \includegraphics[width=0.6\linewidth]{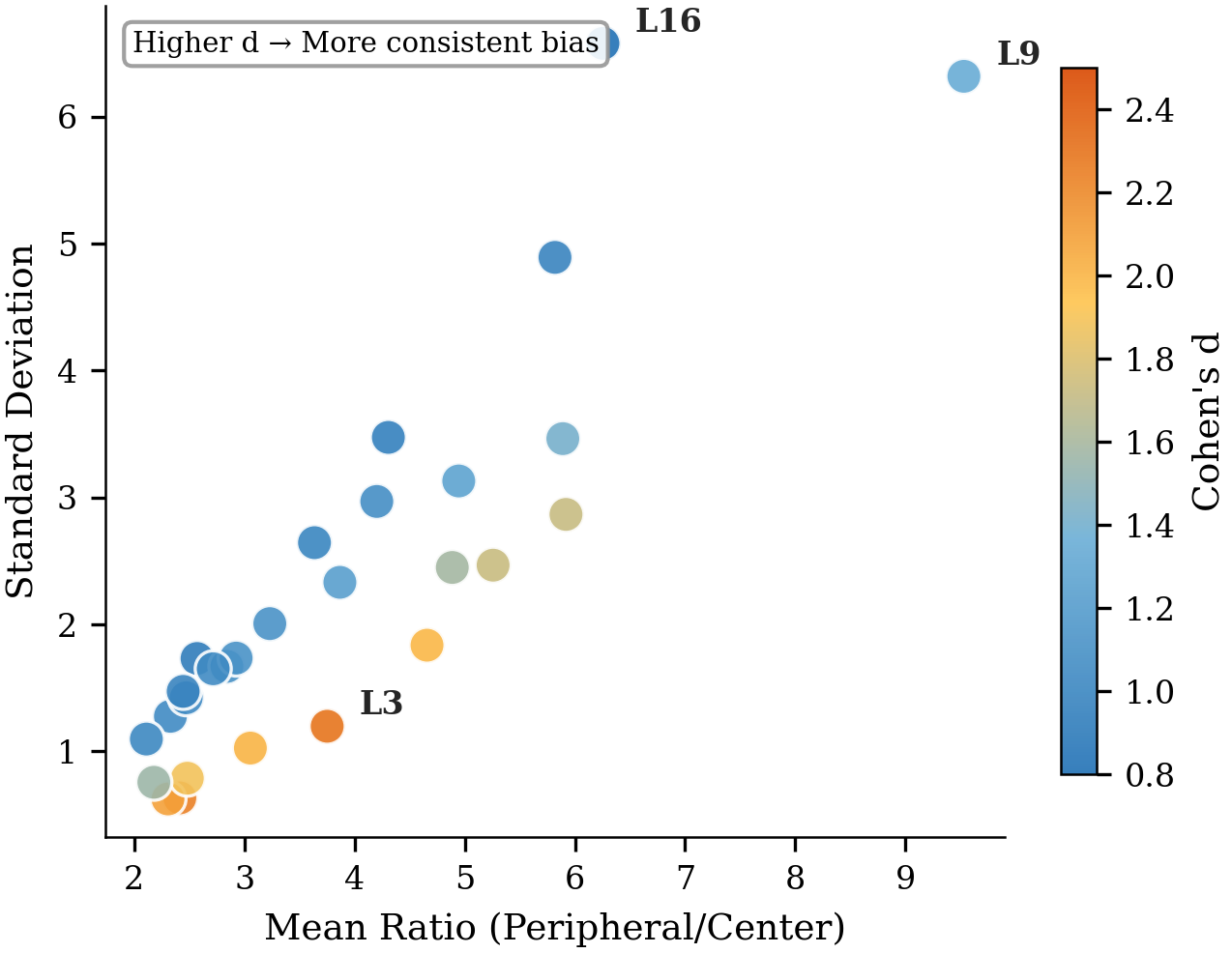}
    \caption{Bias magnitude vs.\ consistency. Cohen's $d = (\mu - 1)/\sigma$ penalizes variance: Layer 9 has peak ratio ($9.53\times$) but moderate $d$ (1.35); Layer 3 achieves highest $d$ (2.30) with lower ratio ($3.75\times$).}
    \label{fig:cohens_d_explanation}
\end{figure}

Figure~\ref{fig:cohens_d_explanation} shows why high bias magnitude does not imply high effect size. Layers with high Cohen's $d$ provide more reliable pruning decisions due to consistent bias patterns.

\subsection{Design Implications}
\label{appendix:design}

\textbf{Pruning Layer Selection.} Effective pruning layers should satisfy two criteria: (1) intervene before bias escalation to prevent error propagation, and (2) exhibit high Cohen's $d$ for reliable decisions. The ablation study empirically validates this strategy.

\textbf{Fusion Weight $\gamma_l$.} Shallow layers exhibit the highest consistency ($d = 1.97$), making event-based calibration most reliable there. We adopt a decay schedule where $\gamma_l$ decreases from shallow to deep layers. The ablation study empirically validates this strategy.

%%%%%%%%%%%%%%%%%%%%%%%%%%%%%%%%%%%%%%%%%%%%%%%%%%%%%%%%%%%%%%%%%%%%%%%%%%%%%%%
% Appendix B: Long-tailed Distribution Analysis
%%%%%%%%%%%%%%%%%%%%%%%%%%%%%%%%%%%%%%%%%%%%%%%%%%%%%%%%%%%%%%%%%%%%%%%%%%%%%%%
\section{Long-tailed Distribution Analysis}
\label{appendix:long_tail}

This appendix analyzes the distributional properties of attention scores and event density, which motivates our rank-based fusion strategy in EARF (Section~\ref{sec:earf}).

\begin{figure}[ht]
    \centering
    \includegraphics[width=0.6\linewidth]{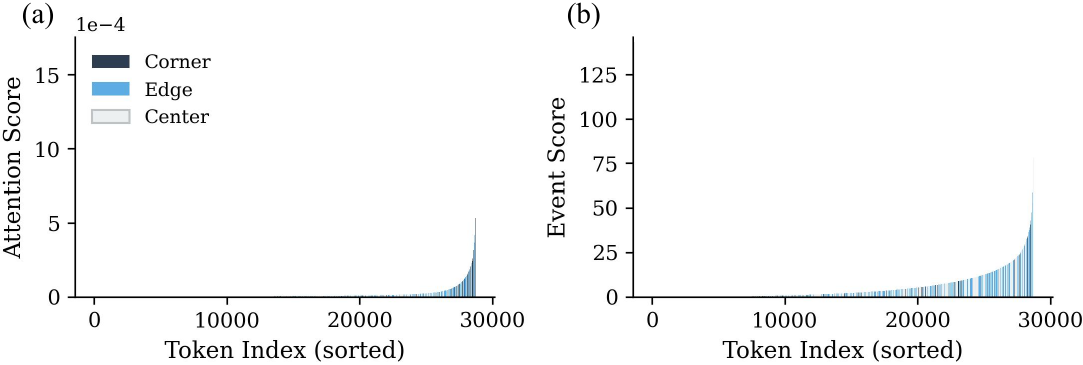}
    \caption{Long-tailed distribution of attention scores and event density. Both modalities exhibit severe skewness, motivating rank-based fusion.}
    \label{fig:long_tail}
\end{figure}

As shown in Figure~\ref{fig:long_tail}, both attention scores and event density exhibit severe long-tailed distributions: a small fraction of tokens dominate the score mass while the majority cluster near zero. This distributional skewness poses a fundamental challenge for multimodal fusion---direct arithmetic combination (e.g., weighted sum) would be dominated by outliers from either modality, failing to capture the relative importance across the full token population. Our rank-based projection normalizes these skewed distributions into uniform ranks, enabling fair comparison and robust fusion.

%%%%%%%%%%%%%%%%%%%%%%%%%%%%%%%%%%%%%%%%%%%%%%%%%%%%%%%%%%%%%%%%%%%%%%%%%%%%%%%
% Appendix C: Mechanistic Analysis of KV-Cache Purification
%%%%%%%%%%%%%%%%%%%%%%%%%%%%%%%%%%%%%%%%%%%%%%%%%%%%%%%%%%%%%%%%%%%%%%%%%%%%%%%
\section{Mechanistic Analysis of KV-Cache Purification}
\label{sec:appendix_C}

This appendix provides an analytical interpretation of how event-unsupported peripheral sink tokens can degrade downstream key--value states in Video-LLMs, and how EARF mitigates this issue through event-assisted rank calibration. We use ``KV cache'' broadly to refer to the active keys and values retained for subsequent attention computation.

\subsection{\texorpdfstring{Problem: Peripheral Sink Tokens in the KV Cache}{Problem: Peripheral Sink Tokens in the KV Cache}}

For clarity, we focus on one pruning layer \(l\). EARF performs score calibration and top-\(K\) pruning within each keyframe, but the retained tokens are then merged into the active sequence for subsequent multimodal self-attention. Let \(\mathcal{T}^{(l)}\) denote the always-retained text and special tokens, and let \(\mathcal{V}^{(l)}=\bigcup_f \mathcal{V}^{(l,f)}\) denote all active visual tokens before pruning, where \(\mathcal{V}^{(l,f)}\) is the active visual-token set of keyframe \(f\). The active key--value set is:
\begin{equation}
{
\mathcal{K}^{(l)}=\mathcal{T}^{(l)}\cup\mathcal{V}^{(l)} .
}
\end{equation}
For a query vector \(q\), attention over the global active set \(\mathcal{K}^{(l)}\) produces:
\begin{equation}
{
    z_q^{(l)}
    =
    \sum_{i \in \mathcal{K}^{(l)}} \alpha_{q,i}^{(l)} v_i^{(l)},
    \quad
    \alpha_{q,i}^{(l)}
    =
    \frac{
    \exp\left(q^\top k_i^{(l)} / \sqrt{d_k}\right)
    }{
    \sum_{j \in \mathcal{K}^{(l)}}
    \exp\left(q^\top k_j^{(l)} / \sqrt{d_k}\right)
    },
}
\end{equation}
where \(d_k\) denotes the key dimension. Pruning changes only the visual-token subset of \(\mathcal{K}^{(l)}\), while text and special tokens are always kept.

As demonstrated in Appendix~\ref{appendix:attention_bias}, visual attention in the evaluated Video-LLM exhibits a consistent peripheral sink: border and corner tokens receive inflated attention scores relative to central tokens. Let \(\mathcal{P}^{(l,f)}\subseteq\mathcal{V}^{(l,f)}\) denote event-unsupported peripheral visual tokens in keyframe \(f\), and let \(\mathcal{P}^{(l)}=\bigcup_f\mathcal{P}^{(l,f)}\). Define \(\mathcal{S}^{(l)}=\mathcal{K}^{(l)}\setminus\mathcal{P}^{(l)}\) as the remaining active tokens. The attention output can be decomposed as:
\begin{equation}
{
    z_q^{(l)}
    =
    \underbrace{
    \sum_{i\in\mathcal{S}^{(l)}} \alpha_{q,i}^{(l)} v_i^{(l)}
    }_{\text{remaining retained-token contribution}}
    +
    \underbrace{
    \sum_{j\in\mathcal{P}^{(l)}} \alpha_{q,j}^{(l)} v_j^{(l)}
    }_{\text{event-unsupported peripheral contribution}} .
}
\end{equation}
When attention-based pruning preferentially retains tokens in \(\mathcal{P}^{(l)}\), these tokens occupy the limited visual-token budget and continue contributing keys and values to later attention computation. Under high compression, each such token also displaces a potentially structure-relevant visual token, weakening the retained visual context.

\subsection{\texorpdfstring{Mechanism: Cascading Degradation Across Layers}{Mechanism: Cascading Degradation Across Layers}}

The degradation can propagate through the network. At pruning layer \(l\), if the retained visual-token set contains many event-unsupported peripheral tokens, their values contribute to the layer output and their representations are forwarded through residual connections and feed-forward blocks. Subsequent attention layers then operate on this biased retained set. If later pruning decisions are again based mainly on attention scores, the same peripheral sink tokens can be repeatedly favored, causing the error to compound.

This mechanism is most harmful at aggressive retention ratios, where the token budget is too small to mask suboptimal token placement. This is consistent with the empirical results in Table~\ref{tab:spatial_results_even}: attention-based baselines such as FastV and PyramidDrop degrade substantially at \(20\%\) retention, whereas ECP maintains robust accuracy by incorporating event-guided calibration.

\subsection{\texorpdfstring{Solution: EARF as a Purification Mechanism}{Solution: EARF as a Purification Mechanism}}

EARF mitigates this issue by using event saliency only at the score-and-mask level, without encoding events as additional hidden states. For pruning layer \(l\) and keyframe \(f\), EARF uses the text-conditioned attention score \(A_i^{(l,f)}\) from the preceding head-averaged post-softmax multimodal self-attention map, following the main-text definition, and the EMSF event saliency \(M_i^{(f)}\) for the same visual token. Although attention is computed over the global active sequence, EARF calibrates and ranks visual tokens within each keyframe set \(\mathcal{V}^{(l,f)}\), matching the implementation. To avoid scale mismatch, both signals are projected into a shared per-keyframe rank space:
\begin{equation}
{
R_{\mathcal{V}^{(l,f)}}(\phi,i)
=
\frac{
\operatorname{rank}_{\mathcal{V}^{(l,f)}}(\phi_i)
}{
\max(|\mathcal{V}^{(l,f)}|-1,1)
},
\quad
\phi\in\{A^{(l,f)},M^{(f)}\}.
}
\end{equation}
The calibrated EARF score is:
\begin{equation}
{
S_{\mathrm{calib}}^{(l,f,i)}
=
(1-\gamma_l)
R_{\mathcal{V}^{(l,f)}}(A^{(l,f)},i)
+
\gamma_l
R_{\mathcal{V}^{(l,f)}}(M^{(f)},i).
}
\end{equation}

Consider an event-unsupported peripheral sink token \(j\) with high attention rank but weak event support:
\[
{
R_{\mathcal{V}^{(l,f)}}(A^{(l,f)},j)\approx 1,
\quad
R_{\mathcal{V}^{(l,f)}}(M^{(f)},j)\approx 0.
}
\]
Attention-only pruning would assign this token a near-maximal score. In EARF, however, its calibrated score becomes:
\begin{equation}
{
S_{\mathrm{calib}}^{(l,f,j)}
\approx
1-\gamma_l.
}
\end{equation}
By contrast, an event-supported token \(u\) with comparable attention rank and high event rank obtains:
\[
{
S_{\mathrm{calib}}^{(l,f,u)}\approx 1.
}
\]
More generally, the score gap between an event-supported token \(u\) and an event-unsupported token \(j\) in the same keyframe is:
\begin{equation}
{
S_{\mathrm{calib}}^{(l,f,u)}
-
S_{\mathrm{calib}}^{(l,f,j)}
=
(1-\gamma_l)\Delta R_A
+
\gamma_l\Delta R_M,
}
\end{equation}
where \(\Delta R_A=R_{\mathcal{V}^{(l,f)}}(A^{(l,f)},u)-R_{\mathcal{V}^{(l,f)}}(A^{(l,f)},j)\) and \(\Delta R_M=R_{\mathcal{V}^{(l,f)}}(M^{(f)},u)-R_{\mathcal{V}^{(l,f)}}(M^{(f)},j)\). When the two tokens have comparable attention ranks and their event ranks differ substantially, the margin is dominated by the event term and is approximately \(\gamma_l\). Importantly, EARF does not simply penalize all peripheral tokens: a peripheral token with strong event support can still be retained. The mechanism specifically targets tokens whose attention rank is inflated but whose event rank is low.

Given the layer-wise retention budget \(K_l^{(f)}\), EARF keeps the visual-token set \(\mathcal{I}_{\mathrm{EARF}}^{(l,f)}\) with the highest calibrated scores in each keyframe. After per-keyframe pruning, the retained visual tokens are merged across keyframes, yielding:
\begin{equation}
{
    \widetilde{\mathcal{V}}^{(l)}
    =
    \bigcup_f \mathcal{I}_{\mathrm{EARF}}^{(l,f)},
    \quad
    \widetilde{\mathcal{K}}^{(l)}
    =
    \mathcal{T}^{(l)}
    \cup
    \widetilde{\mathcal{V}}^{(l)} .
}
\end{equation}
For any pruned visual token \(j\notin\widetilde{\mathcal{V}}^{(l)}\), its key and value are no longer forwarded to subsequent layers. The attention output is therefore computed over a purified active set:
\begin{equation}
{
    \widetilde{z}_q^{(l)}
    =
    \sum_{i\in\widetilde{\mathcal{K}}^{(l)}}
    \widetilde{\alpha}_{q,i}^{(l)} v_i^{(l)},
    \quad
    \widetilde{\alpha}_{q,i}^{(l)}
    =
    \frac{
    \exp\left(q^\top k_i^{(l)} / \sqrt{d_k}\right)
    }{
    \sum_{r\in\widetilde{\mathcal{K}}^{(l)}}
    \exp\left(q^\top k_r^{(l)} / \sqrt{d_k}\right)
    } .
}
\end{equation}
This removes event-unsupported position-biased keys and values from downstream attention, increasing the relative proportion of tokens that carry semantic, motion, or geometric support.

\subsection{\texorpdfstring{Connection to Attention Sink Literature}{Connection to Attention Sink Literature}}

Prior work on efficient LLMs identified that certain tokens in autoregressive models can act as ``attention sinks'', accumulating attention mass regardless of semantic relevance. Our analysis reveals an analogous spatial phenomenon in vision-language models: peripheral visual tokens can serve as two-dimensional attention sinks due to positional biases.

The key difference is that event streams provide an external motion/structure cue for distinguishing structure-supported tokens from event-unsupported peripheral artifacts. Unlike text-only settings where sink tokens must be detected from internal attention statistics alone, EARF leverages token-aligned event saliency as a training-free prior to calibrate pruning decisions. This external prior reduces the downstream presence of event-unsupported position-biased keys and values while preserving semantic attention through the layer-wise weight \(\gamma_l\).

\section{Simulated Event Generation Protocol}
\label{appendix:vid2e_protocol}

For RGB-only public benchmarks, we generate simulated events offline with vid2e from each original source video at its native frame rate. If a video is recorded at 30 fps, vid2e is run on the full 30-fps sequence and synthesizes events between adjacent original frames by triggering events when interpolated logarithmic brightness changes exceed the contrast thresholds. We use the standard setting \(c_p=0.2\), \(c_n=0.2\), and \(t_{\mathrm{ref}}=0\), where \(c_p\) and \(c_n\) are the positive and negative contrast thresholds and \(t_{\mathrm{ref}}\) is the refractory period. This offline synthesis is used only to construct paired RGB-event inputs for benchmarks without native events; it is excluded from the reported online inference latency/GFLOPs. For ESR-Real, events are captured directly by the Prophesee EVK4 hardware and do not require vid2e simulation.

%%%%%%%%%%%%%%%%%%%%%%%%%%%%%%%%%%%%%%%%%%%%%%%%%%%%%%%%%%%%%%%%%%%%%%%%%%%%%%%
% Appendix D: Dataset
%%%%%%%%%%%%%%%%%%%%%%%%%%%%%%%%%%%%%%%%%%%%%%%%%%%%%%%%%%%%%%%%%%%%%%%%%%%%%%%
\section{ESR-Real Dataset}
\label{appendix:esr_real}

% 定义图片中的背景色
\definecolor{recallblue}{HTML}{E1F5FE}
\definecolor{perceptiongreen}{HTML}{E8F5E9}
\definecolor{reasoningred}{HTML}{FFEBEE}
\definecolor{navigationorange}{HTML}{FFF3E0}

To comprehensively evaluate the embodied cognition capabilities of models during motion, we designed a structured benchmark consisting of three primary dimensions: \textit{Perception, Cognition}, and \textit{Reasoning \& Planning}. Each cognitive ability is manifested through specific sub-tasks. We provided handcrafted question prototypes for each task to ensure the evaluation captures the model's spatial reasoning rather than linguistic patterns. The task set logic is as follows:
\begin{itemize}
    \item \textbf{Perception:} This level evaluates the model's foundational ability to recognize environmental goals (Goal Perception) and maintain visual continuity during camera movement (Trajectory Scene).
    \item \textbf{Cognition:} Moving beyond raw vision, these tasks measure the model's capacity to internalize spatial metrics, such as calculating relative proximity (Relative Distance) and predicting metric changes during dynamic actions (Motion Assumption).
    \item \textbf{Reasoning \& Planning: }This is the highest level of embodied intelligence, requiring the model to resolve orientation-based logic (Direction Reasoning) and generate strategic action sequences based on current orientation and target goals (Route Planning).

\end{itemize}

\begin{figure}[ht]
    \centering
    \includegraphics[width=0.9\linewidth]{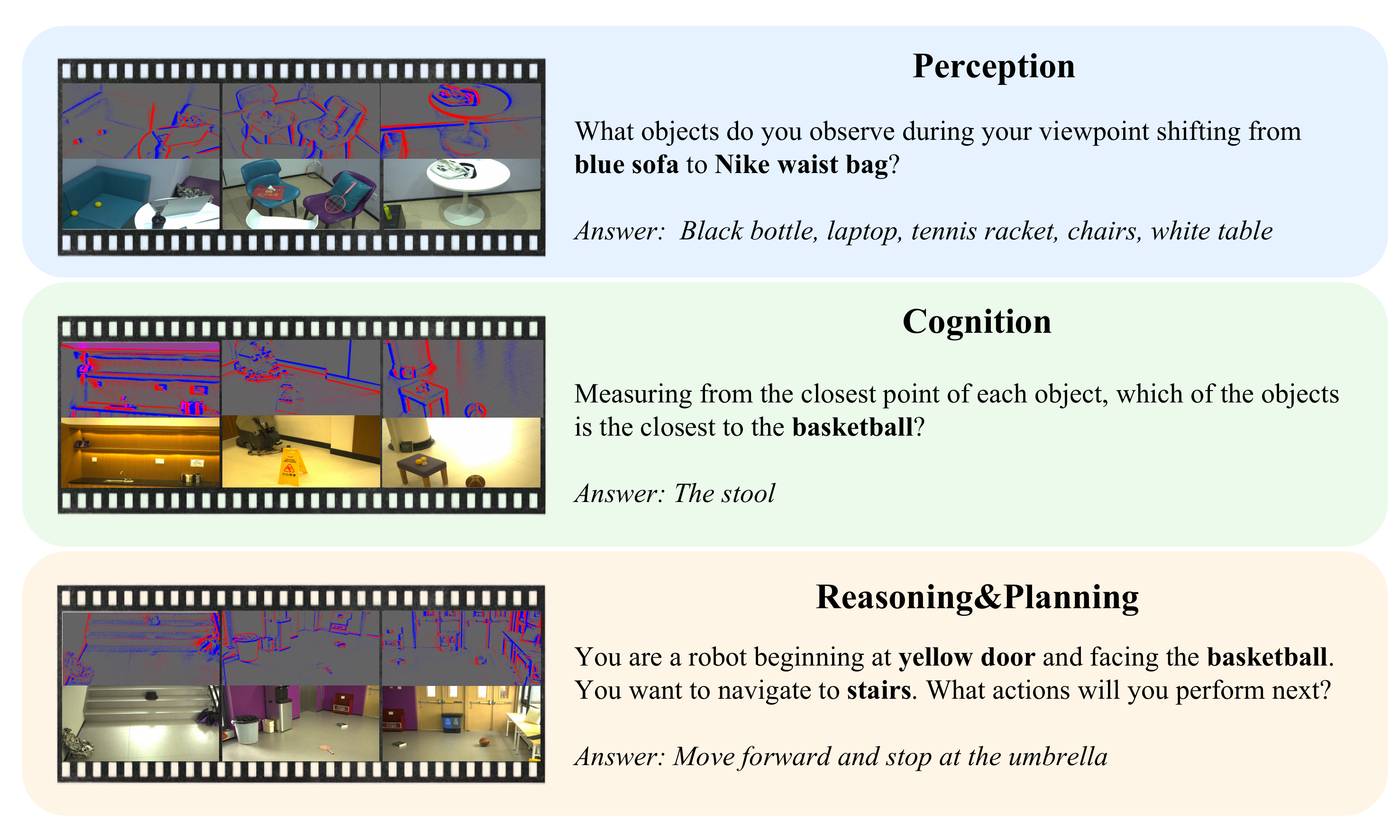}
    \caption{ESR-Real dataset overview}
    \label{fig:dataset_overview_appendix}
\end{figure}

\begin{table}[htbp]
\caption{\textbf{ESR-Real task set overview.} }
\label{tab:task-overview}
\small
\renewcommand{\arraystretch}{1.5} 

\begin{tabularx}{\textwidth}{l X}
%\toprule
% 第一组
\rowcolor{recallblue} \multicolumn{2}{c}{\textbf{Perception}} \\
\textbf{Goal Perception} & At the initial position, you are asked to navigate to the \{target\_object\}. When you reach the last frame of the video, is the destination within view at your current location? \\
\textbf{Trajectory Scene} & What objects do you observe during your viewpoint shifting along \{trajectory\_reference\} \\

% 间距处理
\noalign{\smallskip} 
%  \\[0.5em]

% 第二组
\rowcolor{perceptiongreen} \multicolumn{2}{c}{\textbf{Cognition}} \\
\textbf{Relative Distance} & Measuring from the closest point of each object, which of these objects is the closest to the \{target\_object\}? \\
\textbf{Motion Assumption} & How does the distance between you and the \{target\_object\} change during your \{action\_description\}? \\

\noalign{\smallskip}
%  \\[0.5em]

% 第三组 - 重点修复了下面的 & 符号
\rowcolor{reasoningred} \multicolumn{2}{c}{\textbf{Reasoning \& Planning}} \\
\textbf{Direction Reasoning} & You're a robot standing by the center of \{position\_object\} and facing the center of \{orient\_object\}, is the center of \{query\_object\} to my left, right, or back? (Back defined as $\ge$ 135 degrees turn) \\
\textbf{Route Planning} & You are a robot beginning at \{position\_object\} and facing the \{orient\_object\}. You want to navigate to \{end\_object\}. What actions will  you perform next? \\
%\bottomrule
\end{tabularx}
\end{table}

% 三个比例下的flops和mem
% \begin{table*}[t]
% \caption{Comparison of theoretical computational cost (GFLOPs) and peak memory usage (Mem) on the Urban dataset across varying token retention ratios. \textbf{Ours} consistently minimizes both GFLOPs and memory footprint, outperforming state-of-the-art baselines. The Baseline row indicates the resource usage of the original unpruned model.}
% \label{tab:urban_horizontal}
% \vskip 0.15in
% \begin{center}
% \begin{small}
% \begin{sc}
% \begin{tabular}{l cc c cc c cc}
% \toprule
% \multirow{2}{*}{Method} & \multicolumn{2}{c}{\textbf{20\% Retention}} & & \multicolumn{2}{c}{\textbf{50\% Retention}} & & \multicolumn{2}{c}{\textbf{70\% Retention}} \\
% \cmidrule{2-3} \cmidrule{5-6} \cmidrule{8-9}
%  & GFLOPs $\downarrow$ & Mem (MB) & & GFLOPs $\downarrow$ & Mem (MB) & & GFLOPs $\downarrow$ & Mem (MB) \\
% \midrule
% \textit{Baseline} & \textit{294.6} & \textit{18848} & & \textit{294.6} & \textit{18848} & & \textit{294.6} & \textit{18848} \\
% FastV & 173.9 & 17068 & & 214.3 & 17169 & & 257.8 & 17355 \\
% DivPrune & 187.1 & 17169 & & 214.3 & 17169 & & 257.8 & 17355 \\
% PDrop & 187.5 & 17169 & & 218.8 & 17169 & & 255.0 & 17231 \\
% VTW & 189.8 & 17200 & & 232.7 & 17339 & & 256.5 & 17417 \\
% BTP & 187.5 & 17169 & & 218.8 & 17169 & & 255.0 & 17233 \\
% \midrule
% \textbf{Ours} & \textbf{141.9} & \textbf{16637} & & \textbf{163.3} & \textbf{16664} & & \textbf{179.0} & \textbf{16756} \\
% \bottomrule
% \end{tabular}
% \end{sc}
% \end{small}
% \end{center}
% \vskip -0.1in
% \end{table*}

\subsection{Hardware Setup}
\label{appendix:hardware}

ESR-Real is captured using a synchronized RGB-event camera rig consisting of two sensors rigidly mounted on a custom handheld platform:

\textbf{Event Camera:} Prophesee EVK4 HD, equipped with the Sony IMX636ES event-based vision sensor. Key specifications include: $1280 \times 720$ pixel resolution, $4.86\,\mu\text{m}$ pixel pitch, $>$120\,dB dynamic range, microsecond-level temporal resolution (equivalent to $>$10,000\,fps), and low-light sensitivity down to 0.08\,lux. The asynchronous pixel-level readout enables motion capture without motion blur, making it ideal for first-person dynamic spatial reasoning under ego-motion.

\textbf{RGB Camera:} Hikvision MV-CS050-10UC industrial camera, featuring the Sony IMX264 global shutter CMOS sensor. Specifications: $2448 \times 2048$ pixel resolution (5\,MP), $3.45\,\mu\text{m}$ pixel pitch, up to 60\,fps frame rate, and USB 3.0 interface. The global shutter eliminates rolling shutter artifacts, ensuring geometric consistency with the event stream.

Both cameras are rigidly mounted on a 3D-printed bracket with a baseline of approximately 5\,cm, enabling synchronized capture of the same scene from nearly co-located viewpoints.

\subsection{Stereo Calibration and Alignment}
\label{appendix:calibration}

To project RGB frames and event streams into a common coordinate system, we perform stereo calibration. The calibration procedure consists of three stages:

\textbf{Camera Calibration:} We employ Zhang's flexible calibration method~\citep{zhang2000flexible} for both intrinsic and extrinsic parameter estimation. For the RGB camera, we capture 20--30 images of a $4 \times 7$ checkerboard pattern (square size: 80\,mm) at various orientations and distances. Corner detection is performed using sub-pixel refinement, and the camera parameters are estimated via nonlinear optimization. For the event camera, we use Prophesee's Metavision SDK calibration pipeline with a blinking checkerboard pattern displayed on a monitor. The blinking pattern generates consistent events at pattern boundaries, enabling reliable corner detection directly from the event stream.

\textbf{Stereo Calibration:} To estimate the extrinsic parameters between the RGB and event cameras, both sensors simultaneously capture the same $4 \times 7$ checkerboard pattern. Since event cameras output asynchronous event streams rather than intensity images, we first reconstruct intensity frames from the event stream using E2VID~\citep{rebecq2019e2vid}. Using the detected corners from both the RGB frames and E2VID-reconstructed event frames, we estimate the rotation matrix $\mathbf{R}$ and translation vector $\mathbf{t}$ that define the rigid transformation between the two camera coordinate systems.

\textbf{Event-RGB Alignment:} Using the calibrated stereo parameters, we warp RGB frames into the event camera's coordinate system. Event streams remain in their native resolution ($1280 \times 720$ pixels), while RGB frames are projected onto the event image plane to achieve pixel-level alignment.

\subsection{Data Collection Protocol}
\label{appendix:collection}

Data collection was conducted across diverse indoor environments including offices, laboratories, corridors, and common areas. Each recording session involves handheld traversal with natural ego-motion (walking, turning, approaching objects), with sequence durations ranging from 50 to 100 seconds. RGB video and continuous event streams are recorded simultaneously, followed by post-capture annotation of spatial reasoning QA pairs by the internal research team.

The final ESR-Real benchmark comprises over 700 QA pairs across 6 task categories, covering perception, cognition, and reasoning capabilities essential for embodied AI systems.

\subsection{MCQ Generation}
\label{appendix:mcq_generation}
This section details our automated annotation pipeline, designed to scale the production of high-quality spatial reasoning benchmarks. The objective is to leverage the robust visual perception and logical reasoning capabilities of advanced Large Multimodal Models (LMMs) to transform raw egocentric videos into structured, logic-grounded datasets.

We have designed a multi-stage Chain-of-Thought (CoT) framework tailored for embodied spatial understanding, consisting of the following two primary stages:

\textbf{Stage 1:} \textbf{Translating visual information into a comprehensive topological and spatial metadata map.}

Object Segmentation and Disambiguation: The model identifies the 20 most salient and identifiable objects within the video. To prevent ambiguity, a strict ``Unique Naming'' protocol is enforced. If multiple objects of the same category exist (e.g., two monitors), they are assigned unique identifiers based on objective visual attributes (e.g., ``monitor by the window'') while avoiding unquantifiable subjective adjectives (e.g., ``nice'', ``big'').

Fine-grained Spatial Mapping: The system generates over 25 detailed spatial descriptions covering relative orientations, distances, and overall scene layout.

Proximity Identification: All object pairs within a 1-meter radius are explicitly mapped as ``proximity pairs'' to serve as constraints for subsequent logic generation.

\textbf{Stage 2:} \textbf{Logic-driven QA Synthesis--Populate complex reasoning templates with the structured metadata. }

Spatial Reasoning Logic Constraints: To ensure that questions require high-level reasoning rather than simple local associations, we implement a ``Non-Proximity Constraint.'' For instance, objects selected for Route Planning and Scene Recall must not be proximity pairs, forcing the model to reason over extended trajectories.

Automated Template Population: The system systematically varies reference frames (robot position and orientation) to generate diverse perspectives. Rigorous filters are applied to ensure that the correct options (ground truth) are derived directly from the Stage 1 metadata and that distractors are logically distinct.

By utilizing Gemini-3Pro at this stage, we ensure that the generated questions are grounded in a deep understanding of the 3D environment, providing a challenging and high-fidelity benchmark for embodied AI agents.

\subsection{Manual QA Refinement}
\label{appendix:human_refinement}
The generated MCQs may contain invalid questions, ambiguous options, incorrect answers, and various other issues that require further manual QA refinement. These issues stem from two main sources: a) Even the most advanced Video-LLMs lack the ability to fully understand fine-grained embodied movements within confined indoor spaces. b) The indoor scenarios are densely populated and spatially complex, making dynamic video comprehension challenging. We approach the manual QA refinement process from four aspects:

a) \textbf{Invalid/ambiguous questions: }For example, in Goal Perception tasks, an instruction to ``navigate to the chair'' leads to ambiguity when multiple chairs are present in a living room. The navigation target should be clarified to ``navigate to the wooden armchair near the window'' to ensure precise target identification.

b) \textbf{Object hallucination:} This refers to the presence of furniture or indoor elements in the correct option that were never actually captured during the Trajectory Scene (e.g., mentioning a lamp that was occluded or absent).

c) \textbf{Spatial and metric inaccuracies:} Descriptions involving Direction Reasoning or Relative Distance are often incorrect or imprecise due to the constantly changing camera perspective during motion.

d) \textbf{Choices with insufficient differentiation or logic errors: }In Motion Assumption and Route Planning tasks, one scenario is that distractors are too similar to the correct action sequence. Another is that the ground truth failed to represent the optimal strategic action based on the agent's current orientation.

The entire refinement process required over 200 person-hours. 

\section{Remaining Results of Effectiveness Experiment}
\label{sec:remaining_results}
In the main text, we report six representative task columns from UrbanVideo-Bench and VSI-Bench due to space. This appendix reports the remaining seven task columns; together with Table~\ref{tab:spatial_results_even}, the results cover all 13 selected tasks used to compute Avg.Acc.
Table~\ref{tab:spatial_results_implemental} shows that ECP is particularly stable under aggressive compression. When the retention rate drops to 20\%, VTW decreases to 30.50\%, whereas ECP maintains 37.62\%, close to its 70\% retention result. This trend suggests that event-assisted pruning helps preserve motion- and geometry-relevant evidence when the token budget becomes sparse.
The remaining task columns show the following patterns:

\textbf{a) Perception:} ECP remains strong on Object Recall and Scene Recall. At 20\% retention, it achieves 73.68\% in Object Recall, higher than the full-token baseline's 65.79\%, suggesting that the retained tokens can still cover key environmental entities.

\textbf{b) Cognition:} ECP maintains the highest Cognitive Map accuracy of 54.36\% at 20\% retention, indicating stronger layout consistency under sparse token budgets.

\textbf{c) Reasoning \& Planning:} For High-Level Planning, ECP is consistently strong across pruning ratios, achieving \(50.00\%\) accuracy at \(70\%\) retention and \(52.12\%\) at \(20\%\) retention. At \(70\%\) retention, it clearly outperforms BTP (\(44.49\%\)) and DivPrune (\(42.80\%\)), while other baselines show larger variation across retention ratios.

\begin{table*}[h]
\caption{Accuracy comparison of remaining tasks on first-person dynamic spatial reasoning benchmarks. \textit{Task definitions}: \textbf{Obj.Recall}=Object Recall, \textbf{Scene.Recall}=Scene Recall, \textbf{Abs.Dis}=Absolute Distance Estimation; \textbf{Cog.Map}=Cognitive Map Construction; \textbf{Rel.Dir}=Relative Direction Understanding; \textbf{Route.Plan}=Route Planning;\textbf{ HL.Plan}=High-Level Planning. }
\label{tab:spatial_results_implemental}
\centering

% --- 1. 引入 array 包以支持自定义列宽 (如果导言区已包含可忽略) ---

% --- 2. 定义一个固定宽度的居中列 C ---
% 宽度设为 1.6cm (足够容纳最长表头 Scene.Recall)，Resizebox 会自动缩放整个表格适应页面
\newcolumntype{C}{>{\centering\arraybackslash}p{1.6cm}}

% 消除颜色块上下的白边
\setlength{\aboverulesep}{0pt}
\setlength{\belowrulesep}{0pt}
% 适度调整行高
\renewcommand{\arraystretch}{1.1}

\resizebox{\textwidth}{!}{
% --- 3. 修改列定义：第一列左对齐(l)，剩余8列全部使用等宽列(C) ---
\begin{tabular}{l *{8}{C}}
\toprule

% --- 表头第一行 ---
\multirow{2}{*}{\textbf{Method}} & 
\multirow{2}{*}{\textbf{Avg.Acc}} & 
\multicolumn{2}{c}{\cellcolor{bg_perception}\textbf{Perception}} & 
\multicolumn{2}{c}{\cellcolor{bg_cognition}\textbf{Cognition}} & 
\multicolumn{3}{c}{\cellcolor{bg_planning}\textbf{Reasoning \& Planning}} \\

% --- 表头第二行 ---
 & & Obj.Recall & Scene.Recall & Abs.Dis & Cog.Map & Rel.Dir & Route.Plan & HL.Plan\\
\midrule

% --- Baseline ---
Original       & 36.31 & 65.79 & 69.70 & 23.05 & 51.43 & 34.74 & 29.53 & 45.23 \\
\midrule

% --- 70% Retaining ---
\multicolumn{9}{c}{\textit{70\% Retaining}} \\
\midrule
BTP            & 35.95 & 65.79 & 69.70 & 23.31 & 51.45 & 34.14 & 28.19 & 44.49 \\
DivPrune       & 36.16 & 65.79 & 69.70 & 23.53 & 52.28 & 35.14 & 30.87 & 42.80 \\
FastV          & 36.28 & 65.79 & 69.70 & 23.47 & 51.87 & 34.54 & 28.86 & 44.49 \\
PDrop          & 36.73 & 65.79 & 69.70 & 23.85 & \textbf{53.53} & 35.54 & 30.87 & 44.49 \\
DTD            & 36.36 & 63.16 & 69.70 & 24.01 & 51.45 & 35.74 & 29.53 & 46.61 \\
VTW            & 36.45 & 65.79 & 69.70 & 23.97 & 51.45 & 35.14 & 30.20 & 42.80 \\
DEP            & 36.88 & \textbf{68.40} & 69.70 & 24.28 & 53.50 & \textbf{36.55} & 28.90 & 41.50 \\
\rowcolor{gray!10}OURS          & \textbf{37.78} & 65.79 & \textbf{69.70} & \textbf{24.32} & 52.70 & 33.94 & \textbf{31.54} & \textbf{50.00} \\
\midrule

% --- 50% Retaining ---
\multicolumn{9}{c}{\textit{50\% Retaining}} \\
\midrule
BTP            & 35.09 & 65.79 & 69.70 & 22.20 & 49.38 & 34.14 & 30.20 & 43.64 \\
DivPrune       & 36.77 & 63.16 & 69.70 & 24.05 & 53.53 & \textbf{36.14} & 29.53 & 46.61 \\
FastV          & 35.67 & 65.79 & 66.67 & 22.14 & 50.21 & 34.74 & 30.20 & 47.46 \\
PDrop          & 35.67 & 65.79 & 69.70 & 22.16 & 50.62 & 34.74 & 28.86 & 45.76 \\
DTD            & 36.55 & 63.16 & 69.70 & 24.17 & 53.11 & 35.14 & 29.53 & 47.03 \\
VTW            & 33.84 & 68.42 & \textbf{72.73} & 22.44 & 49.38 & 34.74 & 30.87 & 38.98 \\
DEP            & 33.28 & \textbf{71.10} & 72.70 & 21.18 & 49.00 & 35.14 & 30.20 & 36.50 \\
\rowcolor{gray!10}OURS            & \textbf{38.16} & 63.16 & 69.70 & \textbf{24.50} & \textbf{54.36} & 34.14 & \textbf{32.21} & \textbf{48.73} \\
\midrule

% --- 20% Retaining ---
\multicolumn{9}{c}{\textit{20\% Retaining}} \\
\midrule
BTP            & 33.34 & 65.79 & 66.67 & 19.95 & 51.87 & 34.74 & 30.20 & 47.88 \\
DivPrune       & 34.40 & 60.53 & 66.67 & 20.33 & 49.79 & 35.74 & 27.52 & 51.27 \\
FastV          & 32.64 & 57.89 & 63.64 & 22.79 & 47.72 & 32.73 & \textbf{34.23} & 49.58 \\
PDrop          & 33.12 & 63.16 & 66.67 & 21.50 & 53.53 & 32.53 & 31.54 & 48.31 \\
DTD            & 35.37 & 65.79 & 72.73 & 22.58 & 52.28 & 33.53 & 29.53 & 47.88 \\
VTW            & 30.50 & 42.11 & 54.55 & 12.14 & 47.72 & 35.54 & 30.20 & 41.95 \\
DEP            & 29.95 & 42.10 & 51.50 & 11.11 & 44.10 & \textbf{37.35} & 30.90 & 42.30 \\
\rowcolor{gray!10}OURS          & \textbf{37.62} & \textbf{73.68} & \textbf{72.73} & \textbf{23.05} & \textbf{54.36} & 36.95 & 30.20 & \textbf{52.12} \\
\bottomrule
\end{tabular}
}
\end{table*}

\section{Additional Ablation Study of Component Analysis}
\label{sec:appendix_ablation}
To provide a more granular view of each component, we conduct an ablation study on the combined validation set of UrbanVideo-Bench and VSI-Bench. We analyze Event-Triggered Causal Sampling (ETCS), Event-guided Motion Saliency Filtering (EMSF), and Event-Attention Ranking Fusion (EARF) under token retention rates of 20\%, 50\%, and 70\%, and report averaged accuracy. The results are summarized in Table~\ref{tab:additional_ablation}.

\subsection{Component-wise Contribution Analysis}

\textbf{ETCS: The Spatiotemporal Anchor for Cognition.} Our results empirically validate ETCS as the cornerstone of spatial cognition. As shown in Table~\ref{tab:additional_ablation}, configurations utilizing ETCS  achieve the highest accuracy in Cognition tasks (44.7\%). Comparing ETCS only with EARF only, the inclusion of ETCS yields a significant gain in Cognition (+1.72\%). Without the temporal causality provided by ETCS, downstream reasoning modules operate on misaligned frames, leading to systemic failures in dynamic scene understanding (e.g., mapping and distance estimation).

\textbf{EMSF: The High-SNR Filter for Reasoning.} Unexpectedly, Event-guided Motion Saliency Filtering (EMSF) emerges as the critical driver for logical reasoning. EMSF only achieves the highest performance in Reasoning \& Planning (28.31\%), slightly outperforming even the Full Model. This suggests that EMSF acts as an aggressive denoising mechanism. By filtering out static background redundancies, it maximizes the Signal-to-Noise Ratio (SNR) for dynamic interactions, allowing the LLM to focus purely on causal motion. However, relying solely on EMSF compromises Cognition (43.11\%) due to the loss of background context.

\textbf{EARF: The Semantic Booster for Perception.} Event-Attention Ranking Fusion (EARF) is pivotal for fine-grained visual recognition. Its primary contribution lies in enhancing Perception accuracy. The Full Model (incorporating EARF) achieves the peak Perception score of 50.74\%. Comparing  ETCS+EMSF to (Full), the addition of EARF provides a clear boost in perception reliability (+0.90\%), demonstrating that rank-based fusion effectively harmonizes the high temporal resolution of events with the semantic richness of RGB frames.

\subsection{Synergy and Holistic Superiority of the Full Framework}

While individual modules show strengths in specific sub-domains---ETCS in Cognition and EMSF in Reasoning---Table~\ref{tab:additional_ablation} shows that Full ECP provides the best overall trade-off, with the highest averaged accuracy (37.85\%) and the lowest latency (1.16s) among the tested variants.

\textbf{Why the Full Model  is Optimal.} Despite the specialized strengths of single-module baselines, the Full Model demonstrates superior holistic robustness: \begin{itemize} \item \textbf{Global Maxima:} The Full Model achieves the highest overall accuracy (37.8\%) across all configurations. It effectively harmonizes the aggressive spatial filtering of EMSF with the semantic preservation of EARF. \item \textbf{Optimal Pareto Frontier:} While EMSF alone achieves the highest Reasoning score (28.31\%), it suffers in Cognition and Perception. The Full Model accepts a negligible drop in Reasoning (-0.14\%) to achieve a gain in Perception (+1.54\%) and Cognition (+0.95\%). This trade-off is essential for embodied agents that must simultaneously recognize objects and plan actions. \item \textbf{Synergistic Stability:} The Full Model mitigates the high variance observed in partial configurations. For instance, while ETCS + EARF excels in Perception, it lags behind EMSF in Reasoning accuracy. By integrating all three priors, our framework offers the most balanced and generalizable performance for diverse real-world challenges. 
\end{itemize}

\begin{table*}[t]
\centering
\caption{\textbf{Additional ablation study of component analysis.} We perform a granular breakdown of accuracy metrics mapped to the validation baseline. The best results are \textbf{bolded} and second-best are \underline{underlined}. Full ECP achieves the highest overall accuracy and Perception score while maintaining the lowest latency among multi-component variants.}
\label{tab:additional_ablation}
\vspace{2mm}
\resizebox{1.0\textwidth}{!}{%
\begin{tabular}{lcccccccc}
\toprule
\multirow{2}{*}{\textbf{Variant}} & \multicolumn{3}{c}{\textbf{Components}} & \multirow{2}{*}{\textbf{Latency (s)} $\downarrow$} & \multirow{2}{*}{\textbf{Overall Acc (\%)} $\uparrow$} & \multicolumn{3}{c}{\textbf{Task Categories Acc (\%)}} \\
\cmidrule(lr){2-4} \cmidrule(lr){7-9}
 & \textbf{ETCS} & \textbf{EMSF} & \textbf{EARF} & & & \textbf{Perception} & \textbf{Cognition} & \textbf{R\&P} \\
\midrule
ETCS & \checkmark & - & - & 1.58 & 36.52 & 48.06 & \textbf{44.71} & 26.67 \\
EMSF & - & \checkmark & - & 1.83 & 37.22 & 49.20 & 43.11 & \textbf{28.31} \\
EARF & - & - & \checkmark & 1.53 & 37.15 & 50.25 & 42.99 & 28.18 \\
\midrule
ETCS+EMSF & \checkmark & \checkmark & - & 1.47 & 37.51 & 49.84 & 44.65 & 27.76 \\
ETCS+EARF & \checkmark & - & \checkmark & 1.17 & \underline{37.75} & \underline{50.71} & \underline{44.25} & 28.01 \\
EMSF+EARF & - & \checkmark & \checkmark & 1.59 & 37.00 & 48.66 & 43.68 & 27.93 \\
\midrule
\textbf{Full ECP (Ours)} & \checkmark & \checkmark & \checkmark & \textbf{1.16} & \textbf{37.85} & \textbf{50.74} & 44.06 & \underline{28.17} \\
\bottomrule
\end{tabular}%
}
\end{table*}

\section{Broader Impacts and Responsible Release}
\label{app:broader}

\paragraph{Broader impacts and responsible use.}
ECP reduces the visual-token computation required by Video-LLMs for first-person spatial reasoning, which may lower latency, memory usage, and energy consumption on resource-constrained RGB-event platforms. These efficiency gains may also support more local processing and reduce reliance on cloud offloading. However, as with other egocentric perception technologies, easier deployment in privacy-sensitive environments could raise monitoring or privacy concerns, and incorrect spatial predictions could affect downstream navigation or interaction decisions. ECP should therefore be treated as an efficiency mechanism for perception rather than a complete decision-making system, and safety-relevant deployment requires task-specific validation, privacy-preserving data handling, human oversight, and compliance with applicable regulations.

\paragraph{Responsible release of ESR-Real.}
ESR-Real is collected in controlled indoor environments for spatial-reasoning evaluation, not identity recognition. Before release, we review the RGB-event data, QA annotations, and metadata to remove or blur incidental faces, personal identifiers, sensitive documents or screens, and location-revealing metadata when present. The released benchmark will not include identity, biometric, or person-tracking labels, and will be distributed with documentation and research-use terms restricting re-identification, privacy-invasive monitoring, and unvalidated safety-critical deployment.

% \newpage
% \input{checklist.tex}

\end{document}